\pdfoutput=1

\documentclass[11pt]{article}

\usepackage[final]{acl}

\usepackage{times}
\usepackage{latexsym}

\usepackage[T1]{fontenc}

\usepackage[utf8]{inputenc}

\usepackage{microtype}

\usepackage{inconsolata}

\usepackage{graphicx}
\usepackage{amsfonts}
\usepackage{amsmath}
\usepackage{amssymb}       
\usepackage{bold-extra}    
\usepackage{bm}            
\usepackage{multirow}      
\usepackage{booktabs}      
\usepackage{enumitem}      
\usepackage{subcaption}    
\usepackage{tikz}
\usepackage{pgfplots}
\usepackage[hang,flushmargin]{footmisc}  
\usepackage{listings}
\usepackage{xcolor}

\definecolor{codegray}{gray}{0.9}
\lstdefinestyle{jsonstyle}{
    backgroundcolor=\color{codegray},   
    basicstyle=\ttfamily\footnotesize,
    breakatwhitespace=false,         
    breaklines=true,                 
    captionpos=b,                    
    keepspaces=true,                 
    numbers=none,                    
    showspaces=false,                
    showstringspaces=false,
    showtabs=false,                  
    tabsize=2
}

\title{Do Retrieval-Augmented Language Models Adapt to Varying User Needs?}

{\author{Peilin Wu\textsuperscript{1}\thanks{Equal contribution}, Xinlu Zhang\textsuperscript{2}\footnotemark[1], Wenhao Yu\textsuperscript{3}, Xingyu Liu\textsuperscript{4}, Xinya Du\textsuperscript{1}, Zhiyu Zoey Chen\textsuperscript{1} \\
\textsuperscript{1}Department of Computer Science, The University of Texas at Dallas,\\
\textsuperscript{2}Department of Computer Science, University of California, Santa Barbara,\\
\textsuperscript{3}Tencent AI Seattle Lab,\textsuperscript{4} 
Harvard University\\
\texttt{\{peilin.wu, zhiyu.chen2\}@utdallas.edu}
}}

\begin{document}
\maketitle
\begin{abstract}
Recent advancements in Retrieval-Augmented Language Models (RALMs) have demonstrated their efficacy in knowledge-intensive tasks. However, existing benchmarks often assume a singular view of optimal information use, neglecting diverse user needs where 'correctness' can mean faithfulness to instructed sources over factual recall. This paper introduces a novel evaluation framework that systematically assesses RALMs under three user need cases—Context-Exclusive, Context-First, and Memory-First—across three distinct context settings: Context Matching, Knowledge Conflict, and Information Irrelevant. By varying both user instructions and the nature of retrieved information, our approach captures the complexities of real-world applications where models must adapt to diverse user requirements. Through extensive experiments on multiple QA datasets, including HotpotQA, DisentQA, and our synthetic URAQ dataset, we find that restricting memory usage improves robustness in adversarial retrieval conditions but decreases peak performance with ideal retrieval results and model family dominates behavioral differences. Our findings highlight the necessity of user-centric evaluations in the development of retrieval-augmented systems and provide insights into optimizing model performance across varied retrieval contexts, explicitly separating factual correctness from faithfulness‑to‑instruction so readers know which dimension each score reflects. We will release our code and URAQ dataset upon acceptance of the paper.

\end{abstract}
\section{Introduction}
Recent advances in Language Models (LMs) have yielded impressive performance in knowledge‐intensive tasks through Retrieval Augmented Generation (RAG) \cite{10.5555/3495724.3496517}, including Real‐time Question Answering \cite{wang-etal-2024-rear}, Educational Tutoring \cite{han2024improvingassessmenttutoringpractices}, and Personal Assistants \cite{wang-etal-2024-crafting}. While these applications showcase RAG’s versatility, they also demand LMs that can adapt to diverse user needs—expressed via instructions on whether to prioritize external evidence or internal knowledge, or adhere strictly to specified (even potentially counter-factual) contexts, as seen in compliance, creative writing, or hypothetical scenarios. For instance, Real‐time QA may rely heavily on updated external facts, whereas tutoring may draw more on the model’s conceptual understanding. Despite this potential, current RAG methods still struggle with identifying relevant references \cite{laban2024summaryhaystackchallengelongcontext}, resolving knowledge conflicts \cite{wang2024astuteragovercomingimperfect}, and reasoning effectively \cite{islam-etal-2024-open}. These challenges underscore the need for robust evaluation strategies capturing how well Retrieval Augmented Language Models (RALMs) adapt to evolving user requirements.
\begin{figure}[t]
\centering
\includegraphics[width=\columnwidth]{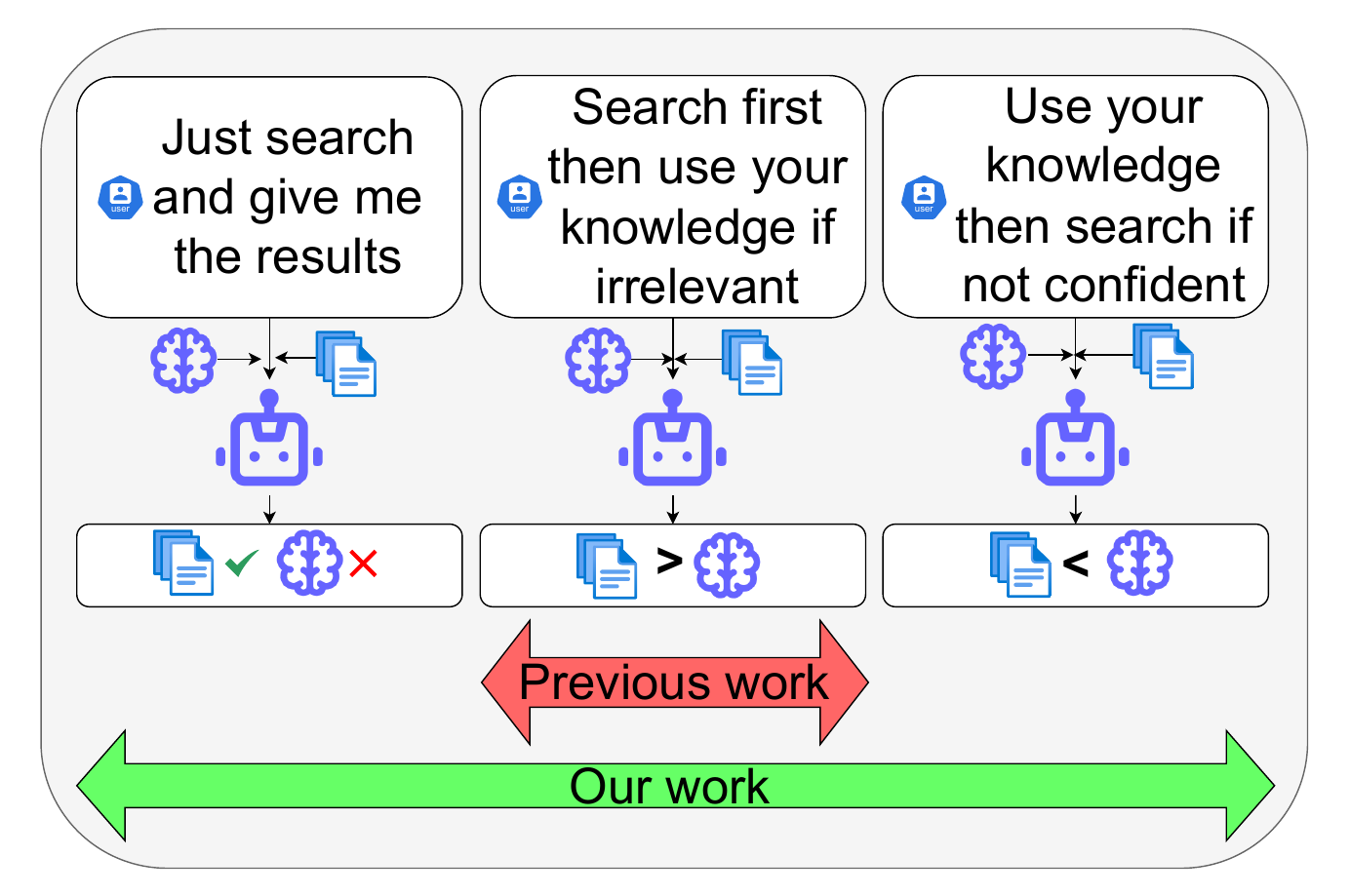}
\caption{User needs may have different directions on how to use retrieved context and internal memory as knowledge sources and most of the previous work only focused on a small portion of them.}
\vspace{-0.2in}
\label{fig:motivation}
\end{figure}

Even though existing RAG/RALM benchmarks \citep{yu2024evaluationretrievalaugmentedgenerationsurvey, es2023ragasautomatedevaluationretrieval, 10.1609/aaai.v38i16.29728}—including those that focus on multi‐scenario evaluations \citep{friel2024ragbenchexplainablebenchmarkretrievalaugmented, zhu2024ragevalscenariospecificrag}—have advanced retrieval‐augmented evaluation, they typically assume a single “optimal” approach to external information (e.g., always relying on retrieved context). This narrow perspective overlooks how diverse user instructions can dramatically alter the desired model's behavior within the same scenario. In medical fact‐checking, for instance, one user might demand answers derived only from peer‐reviewed studies, while another relies on the model’s internal knowledge—even if these sources conflict \citep{Miao2024IntegratingRG}. Such constraints underscore an urgent question: \textit{how can we systematically evaluate LMs under varying context usage requirements to reflect different user needs?}

In this paper, we present a simple yet effective  \emph{evaluation framework} that rigorously examines how Retrieval‐Augmented Language Models (RALMs) respond and be faithful to varying user instructions and context conditions. We consider three generic \textbf{user cases}—\textbf{(1)~Context-Exclusive }, \textbf{(2)~Context-First}, and \textbf{(3)~Memory-First }—to capture different degrees of reliance on external information versus internal knowledge. Alongside these cases, we vary the \textbf{context settings}—\textbf{(a)~Context Matching}, \textbf{(b)~Knowledge Conflict}, and \textbf{(c)~Information Irrelevant}—to represent scenarios where retrieved materials may align with, contradict, or fail to address the query. By intersecting user cases with distinct context conditions, we more closely mirror the complexities of real‐world applications, where both the user’s priorities and the reliability of retrieved information can shift dramatically. This approach reveals how each scenario might alter the correct response—especially when context and memory conflict—an aspect often overlooked in previous work.

We conduct extensive experiments on our curated dataset, URAQ, along with two public datasets, DisentQA \cite{neeman-etal-2023-disentqa} and HotpotQA \cite{yang-etal-2018-hotpotqa}, evaluating two model families, Llama3.1 \citealt{grattafiori2024llama3herdmodels} and Qwen2.5 \citealt{qwen2025qwen25technicalreport}, across various model sizes and numbers of retrieved contexts. Our findings reveal that:  
1) \textbf{Current LMs struggle to satisfy diverse user needs}, achieving below 50\% accuracy across all datasets, with Llama-3.1-8B-Instruct occasionally nearing 0\%. 
2)  \textbf{Contextual restriction alters performance}: Restricting models to rely solely on retrieved context improves LMs performance when external context content is different from internal memory  by up to 23\% accuracy difference on the same model but decreases the performance under ideal retrieval by up to 17\%. 3) \textbf{Model family dominate behavioral differences}: Model family contributes the majority of behavioral differences, which further emphasize the importance of choosing the correct model for different user needs through proper evaluations. For instance, under retrieval with knowledge conflict, Llama3.1 models exhibit a performance decline of up to 10.2\% in accuracy when transitioning from Context-First and Memory-First to the Context-Exclusive case, whereas Qwen2.5 models show the opposite pattern, with an improvement of nearly 20\%.


\section{Related Work}
Our work intersects with four key research areas: (1) Retrieval-Augmented Generation Systems (\S\ref{subsec:recent_rag_systems}), (2) Knowledge Conflict Resolution (\S\ref{subsec:knowledge_conflict}), and (3) RAG Evaluation Benchmarks (\S\ref{subsec:recent_rag_benchmark}). We situate our framework within this landscape and highlight critical gaps in current approaches.

\subsection{ RAG Systems}
\label{subsec:recent_rag_systems}
Modern RAG systems built on foundational architectures like REALM \cite{10.5555/3524938.3525306} and DPR \cite{karpukhin-etal-2020-dense}, which first demonstrated the value of integrating neural retrieval with language modeling. Subsequent work improved context utilization through better attention mechanisms (RETRO \cite{Borgeaud2021ImprovingLM}) and multi-stage reasoning (Atlas \cite{10.5555/3648699.3648950}). While these systems demonstrate impressive performance on knowledge-intensive tasks, they primarily optimize for single objective functions under the implicit assumption that retrieved context should always be prioritized. Recent work on controllable generation (\citealt{li-etal-2023-large}; \citealt{ashok2024controllabletextgenerationinstructiontuning}; \citealt{wei2024instructraginstructingretrievalaugmentedgeneration}) begins to address this limitation but focuses on content style rather than source prioritization. We aim to raise the attention to diversified objectives of RAG system by this work about evaluating performance under different \emph{user needs}.

\subsection{Knowledge Conflict}
\label{subsec:knowledge_conflict}
The challenge of resolving conflicts between internal knowledge and external context has gained attention as LMs and RAG systems mature \cite{xu-etal-2024-knowledge-conflicts}. Early work by \citet{longpre-etal-2021-entity} identified context-memory conflicts as a key failure mode of LMs through evaluation on QA dataset. Subsequent works proposed multiple solutions, including but not limit to various fine-tuning, prompting, or decoding methods, to context-memory conflicts that require LM to be faithful to context in order to ignore outdated knowledge (\citealp{shi-etal-2024-trusting}; \citealp{zhou-etal-2023-context}) or faithful to memory in order to discriminate misinformation are rarely explored \cite{xu-etal-2024-earth}. However, the hybrid strategies that utilize both context and memory with prioritization, although commonly appeared in real-world applications, are rarely explored. In addition, there also exists applications that require LMs and RAG systems to work along or accept fictitious information or knowledge, which are commonly ignored by the previous works. Our framework includes the hybrid strategies that stem from the fundamental \emph{user needs}, providing a wider coverage of evaluating RALMs performance under context-memory conflict situations.

\subsection{Recent RAG Benchmark}
\label{subsec:recent_rag_benchmark}
Previous RAG benchmarks like RAGAS \cite{es2023ragasautomatedevaluationretrieval} and RGB \cite{10.1609/aaai.v38i16.29728} have facilitated progress by quantifying performance across various scenarios. However, many of these benchmarks focused on a single type of optimal setting in terms of context usages (for instance, always prioritizing the context), overlooking how different user instructions may drastically affect model behaviors and performances. Moreover, previous multi-scenario evaluations (\citealt{friel2024ragbenchexplainablebenchmarkretrievalaugmented}; \citealt{zhu2024ragevalscenariospecificrag}), while covering a wide range of specific tasks and purpose abundant metrics for evaluating different aspects of RAG systems, also tend to follow the paradigm of focusing on singular optimality, neglecting that different user needs can actually happen in the same scenario, ultimately hindering the comprehensiveness of benchmark. FaithEval \cite{ming2025faithevallanguagemodelstay} proposes a benchmark to evaluate the faithfulness of the RAG system. Our work diverges by decoupling evaluation criteria from predefined singular optimality and measuring model capability to \textit{adapt} to dynamic \emph{user needs} by using different instructions. This mirrors real-world deployments where systems must honor diverse users' requirements rather than optimize for monolithic accuracy.

\section{Evaluation Framework}
\label{sec:framework}


In this section, we present our evaluation framework to measure Language Models' (LMs') performance. Specifically, we first describe the design of three abstract \textbf{user need cases} (\S\ref{subsec:user_cases}) representing different typical \textit{user needs} expressed by context usages. Then, we describe the three \textbf{context settings} (\S\ref{subsec:context_settings}) motivated by practical usage conditions in which the relevancy of the context varies and may conflict with the LMs' memory.

\subsection{User Need Cases}
\label{subsec:user_cases}
\begin{figure}[t]
\centering
\includegraphics[width=\columnwidth]{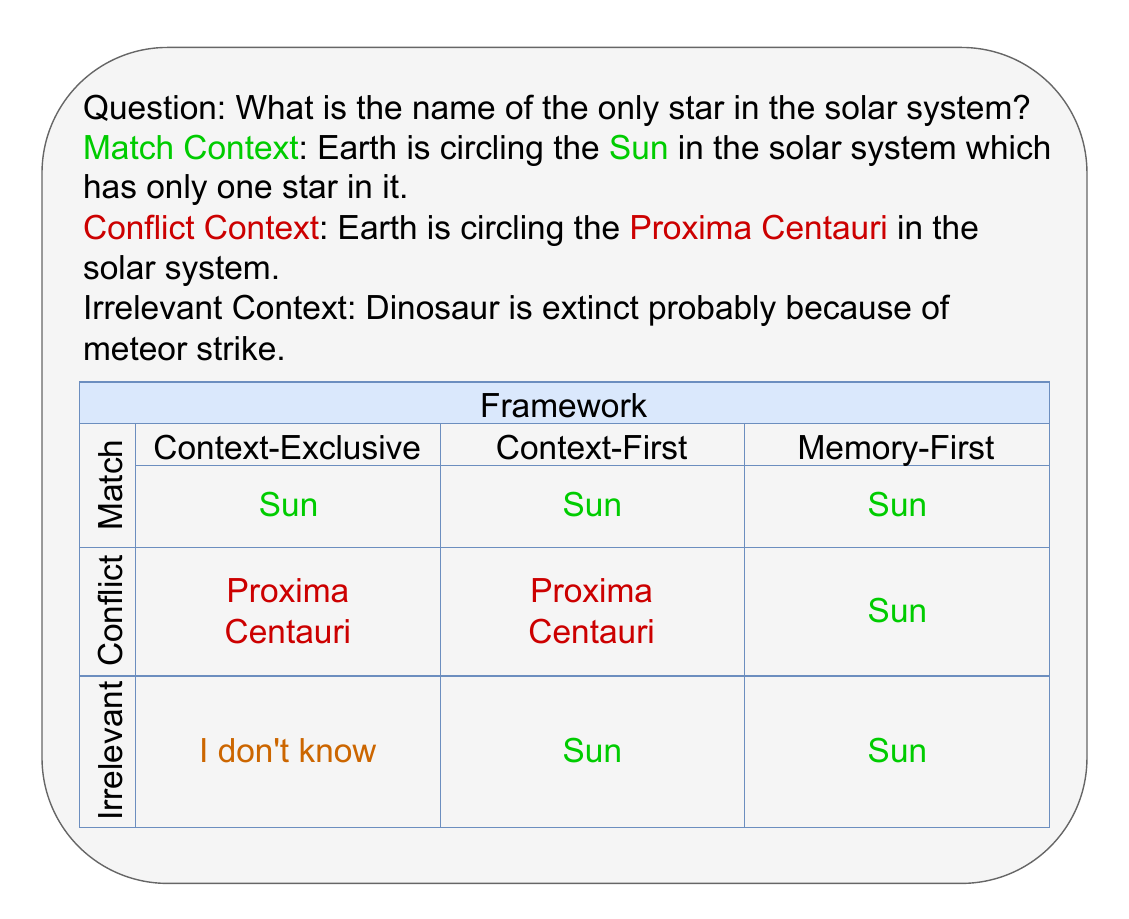}
\vspace{-0.3in}
\caption{An illustration of the framework with an example question with its possible retrieved context and the ground truth answer under each situation. According to different user needs and context settings, the ground truth answer can be different, reflecting instructed faithfulness (e.g., to 'Proxima Centauri' if dictated by context and user need) rather than absolute factual correctness.}
\vspace{-0.2in}
\label{fig:framework}
\end{figure}

To evaluate RALMs under varying \textit{user needs}, we define a spectrum based on reliance on contextual information versus internal memory. This spectrum, illustrated in Figure~\ref{fig:framework}, consists of three distinct \textbf{user needs}, determined by how LMs are instructed. Example prompts are in Appendix~\ref{sec:example_user_need}.

 \paragraph{Context-Exclusive:} LMs must strictly base answers on retrieved context, responding ``I don’t know'' if context is unhelpful. Prompts enforce unconditional adherence to external evidence, eliminating reliance on internal knowledge.
 \paragraph{Context-First:} LMs prioritize retrieved context but fall back on memory when no relevant context exists. Prompts establish context as primary, with memory as a secondary source.
 \paragraph{Memory-First:} LMs rely on internal memory unless uncertain, in which case they defer to retrieved context. Prompts invert the hierarchy, making memory the default unless confidence is low.

\subsection{Context Settings}
\label{subsec:context_settings}
To better analyze RALMs under real-world situations with sub-optimal retrieval results, it is beneficial to also consider the spectrum of context quality on top of each user case. For any context retrieved in an RAG system, we can assess its quality based on two primary dimensions: 1) \textbf{Relevance to the Task or Question}: Whether the retrieved context contains information that is semantically or factually related to the question. 2) \textbf{Alignment with LM’s Internal Knowledge}: Whether the retrieved context supports or contradicts the knowledge that the model already possesses. These two dimensions create a 2 × 2 space (relevant/irrelevant × match/conflict), but due to the nature of irrelevant context (which neither supports nor contradicts), the space reduces to three distinct context settings.

\paragraph{Conext Matching.}
There is at least one retrieved context \emph{relevant} to the question and \emph{matches} with the LM’s memory. This is an ideal situation for RALMs as correct knowledge is presented in both the external context and the internal memory.  

\paragraph{Knowledge Conflict.}
There is at least one retrieved context \emph{relevant} to the question but \emph{conflicts} with the LM’s memory. This setting simulates context-memory knowledge conflicts \cite{xu-etal-2024-knowledge-conflicts} and tests the model’s ability on generation with strictly following instructions regarding context usages.

\paragraph{Information Irrelevant.}
All retrieved contexts are unrelated to the question. This setting simulates the Needle-In-a-Haystack \cite{laban2024summaryhaystackchallengelongcontext} situation and tests the model’s ability on knowledge selection. 

\section{ Experimental Setup}

\label{subsec:datasets}

\subsection{Datasets}
\paragraph{Overview of QA Datasets}  This experiment employs three QA datasets: HotpotQA \cite{yang-etal-2018-hotpotqa}, DisentQA \cite{neeman-etal-2023-disentqa}, and our synthetic User-focused Retrieval-Augmented QA (URAQ). To assess RALMs' real-world performance, we use HotpotQA and DisentQA versions augmented with conflicted knowledge by \citet{shaier-etal-2024-adaptive} for the retrieval-content knowledge conflict setting. While valuable, these benchmarks lack controlled knowledge boundaries and have varying question difficulty, limiting evaluation. They also rely on long-document contexts only, thereby restricting retrieval diversity in terms of document length. In addition, the nature of factual-based for these datasets makes them may not be fully aligned with the evaluation of under needs. URAQ complements these by providing uniformly difficult questions and numerous concise modified contexts, specifically to isolate instruction-following and conflict-resolution capabilities when adapting to varied user needs, distinct from general comprehension over long or highly complex factual texts. While specialized domain datasets (e.g., medical, real-time QA) would be ideal for demonstrating our three user needs, we opted for these known and synthetic benchmarks to ensure reproducibility, broader comparability, and generalizability within budget constraints. The framework itself remains applicable to more domain-specific evaluations.

\begin{table}[htbp!]
\centering\resizebox{\columnwidth}{!}{
\begin{tabular}{c|ccc} 
\toprule
\bf Dataset & \bf Num. of Context Sequence & \bf Size & \bf Max. Token \\
\midrule
Synthetic & 1, 10, 25, 50, 100, 250, 500, 1000 & 231 & 25k \\
DisentQA & 1, 2, 4, 8, 16, 32, 64 & 1415 & 59k \\
HotpotQA & 1, 2, 4, 8, 16, 32 & 1274 & 35k \\
\bottomrule
\end{tabular}}

\caption{Basic information of the three datasets used in the experiment. The number of retrieved context is increased in a exponential way until the average number of tokens at the highest number of each sequence reaches around 20k in order to balance the effectiveness of the experiment on long context and the consumption of computational resources. The number of maximum tokens among all samples for a dataset may vary based on context retrieved.}
\vspace{-0.2in}
\label{tab:dataset_info}
\end{table}

\paragraph{URAQ Construction}  We construct URAQ by first generating simple, distinct knowledge statements via GPT-4o-mini \cite{openai2024gpt4ocard} and removing near-duplicates using SentenceBERT \cite{reimers-gurevych-2019-sentence}, then creating both original and “manipulated” versions by substituting key information or adding negations.  For each knowledge pair, we produce a question requiring 1–5 reasoning steps and two separate answers (one from the original knowledge, one from the manipulated), ultimately selecting the 4-hop subset for the final dataset. A detailed description of this procedure is provided in Appendix \ref{sec:dataset_construction}. To ensure fairness in evaluating multiple models, which may possess different internal knowledge, our experiments (particularly with URAQ) utilize a subset of questions for which the underlying correct factual knowledge is confirmed to be known by all evaluated models. This is achieved by pre-screening models on single-hop versions of questions related to the original knowledge. This pipeline ensures applicability across various domains and enables users to convert any datasets that previously designed for factuality and truthfulness into a faithfulness-oriented dataset, which adapts to our experiment on user needs.

\subsection{Retrieval Context Setup}
To examine how performance changes with varying amounts of retrieved context, rather than using a fixed retrieval count as in previous work \cite{zhu2024ragevalscenariospecificrag}, we evaluate LM performance by exponentially increasing the retrieval count across different datasets, shown in Table \ref{tab:dataset_info}. To assess the models’ tolerance to distracting or irrelevant contexts, we ensure that only one relevant context is present for both the context-matching and conflicting settings, randomly positioned within the prompt. A detailed description of the prompt formatting and example is in the Appendix \ref{sec:example_input_prompt}. All other contexts are selected from a pool of \emph{original} and \emph{manipulated} knowledge that excludes any information directly related to the current question.

\subsection{Evaluation  Metrics}
 To rigorously assess user-need awareness across different user needs with different retrieval content, we test each user need with identical questions but varying the guidance on context usage, spanning three levels:

\paragraph {1. Overall User Need Accuracy}: The model must satisfy \textit{all user needs} simultaneously. Specifically, each test sample can be counted as correct if and only if the model can answer the same question under \textit{all user cases and all context settings}. In this way, we can evaluate the LMs in a generic setting.

\paragraph {2. Case-Level Accuracy}
For each individual user need, we assess the model’s performance across multiple context settings. A test sample is considered correct only if the model consistently provides the correct answer \textit{across all variations of context under that specific user need}. This evaluation method ensures that the model demonstrates reliability in addressing a given requirement, independent of the context variations presented.

\paragraph {3. Setting-Specific Accuracy} In each context setting, test sample is considered correct if the model obtain the answer is same as the ground truth in the corresponding setting. 
By evaluating models at these three levels, we obtain a comprehensive view of how consistently and robustly they meet each user need across different contextual requirements.

\subsection{Evaluation model }
To evaluate user‐need awareness, we conduct comprehensive experiments on 4 Instruct LMs using two distinct open-source LLM families—Llama 3.1 \cite{grattafiori2024llama3herdmodels}, and Qwen 2.5 \cite{qwen2025qwen25technicalreport}—which vary in model size. We set the maximum context length to 128k, the temperature to 0, and Top‐p to 1, while leaving all other configurations at their default values which defers to the Appendix D.
\begin{figure*}[t!]
\centering
\includegraphics[width=\textwidth]{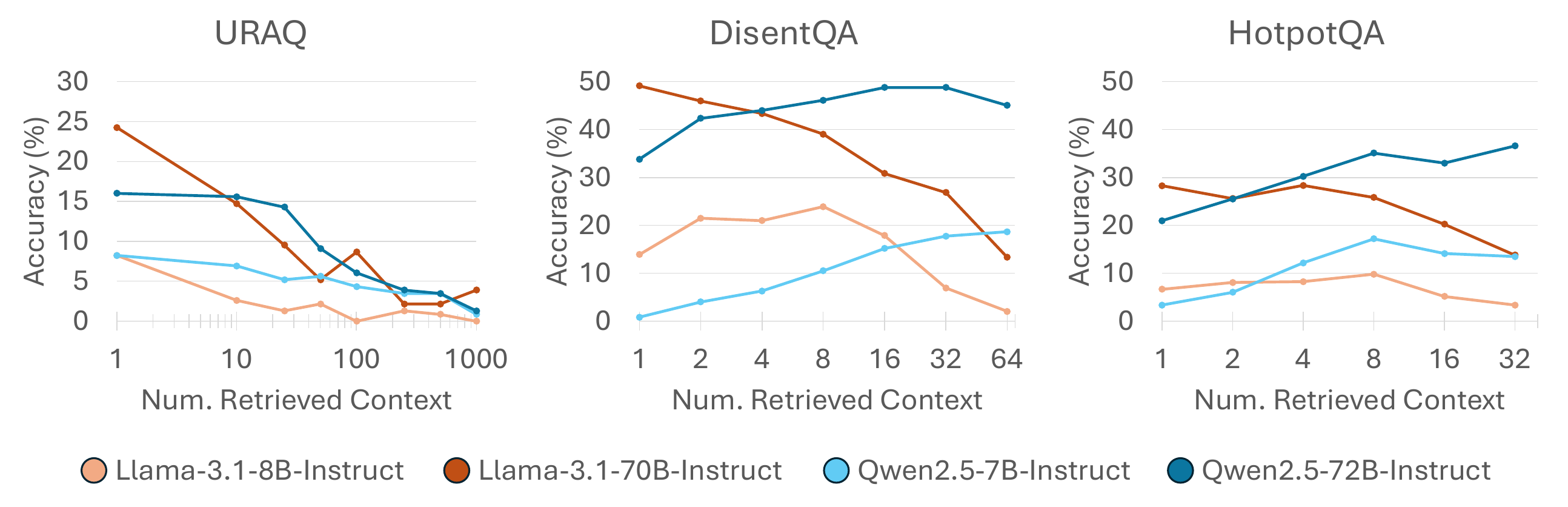}
\vspace{-0.3in}
\caption{Overall user need performance curve of all models on each dataset.}
\vspace{-0.2in}
\label{fig:overall_performance}
\end{figure*}

\section{Result \& Analysis}

\subsection{Overall Performance}

We start our analysis on the overall performance across all three user cases by using the overall user need accuracy to access the capacity of  user need awareness on different LMs. The results are shown in Figure \ref{fig:overall_performance}.


\paragraph{LMs struggle across all datasets, and URAQ is more challenging than existing benchmarks} 
No model surpasses 50\% accuracy across different user needs, with Llama-3.1-8B-Instruct performing particularly poorly, nearing 0\%. While performance is low across all datasets, URAQ proves significantly more challenging than DisentQA and HotpotQA. The best-performing model, Qwen2.5-72B-Instruct, scores up to 44.4\% lower on URAQ. URAQ’s diverse external information, multi-step reasoning, and conflicting knowledge make retrieval and synthesis more challenging for LLMs, emphasizing the need for stronger reasoning capabilities to handle complex real-world user needs.

\paragraph{LMs behave differently at the model-family level but similarly within the same family.} Overall, we observe distinct patterns in LMs across different model families on two out of three datasets. Specifically, there is a clear divergence in behavior between the Qwen2.5 and Llama-3.1 model families on DisentQA and HotpotQA. The Qwen2.5-7B-Instruct and its larger 72B variant exhibit an increasing trend in accuracy as the number of retrieved contexts grows, whereas the Llama-3.1-8B-Instruct and 70B-Instruct models follow a decreasing trend. This difference likely stems from model-specific behavioral tendencies and a potential trade-off between instruction-following capability and multi-hop reasoning ability, which we further discuss in Section \ref{subsec:general_performance_nonideal}. On URAQ, although both model families exhibit declining trends, the Llama-3.1 models experience a steeper drop in performance compared to the Qwen2.5 models. For example, the performance gap from 1 to 10 retrieved contexts in the Qwen family is around relative accuracy 1.5\%, whereas for the Llama-3.1 family, it is 9.1\%, indicating a more pronounced decline.

\paragraph{Larger models exhibit better user needs awareness.} Within the same model family, larger models (70B+/72B) consistently outperform their smaller counterparts (7B/8B), demonstrating improved user needs awareness. Notably, Qwen models exhibit up to a 37.7\% accuracy improvement, while Llama models achieve a 36.3\% gain on DisentQA, highlighting the substantial benefits of scaling model size. However, it is also important to note that the magnitude of performance improvement diminishes as the number of retrieved contexts increases, suggesting potential saturation effects or increased difficulty in effectively leveraging larger context windows.

\subsection{General Performance for Each User Need}
\label{subsec:general_performance_nonideal}

\begin{figure}[t!]
\centering

\begin{subfigure}{\columnwidth}
\centering
\includegraphics[width=\columnwidth]{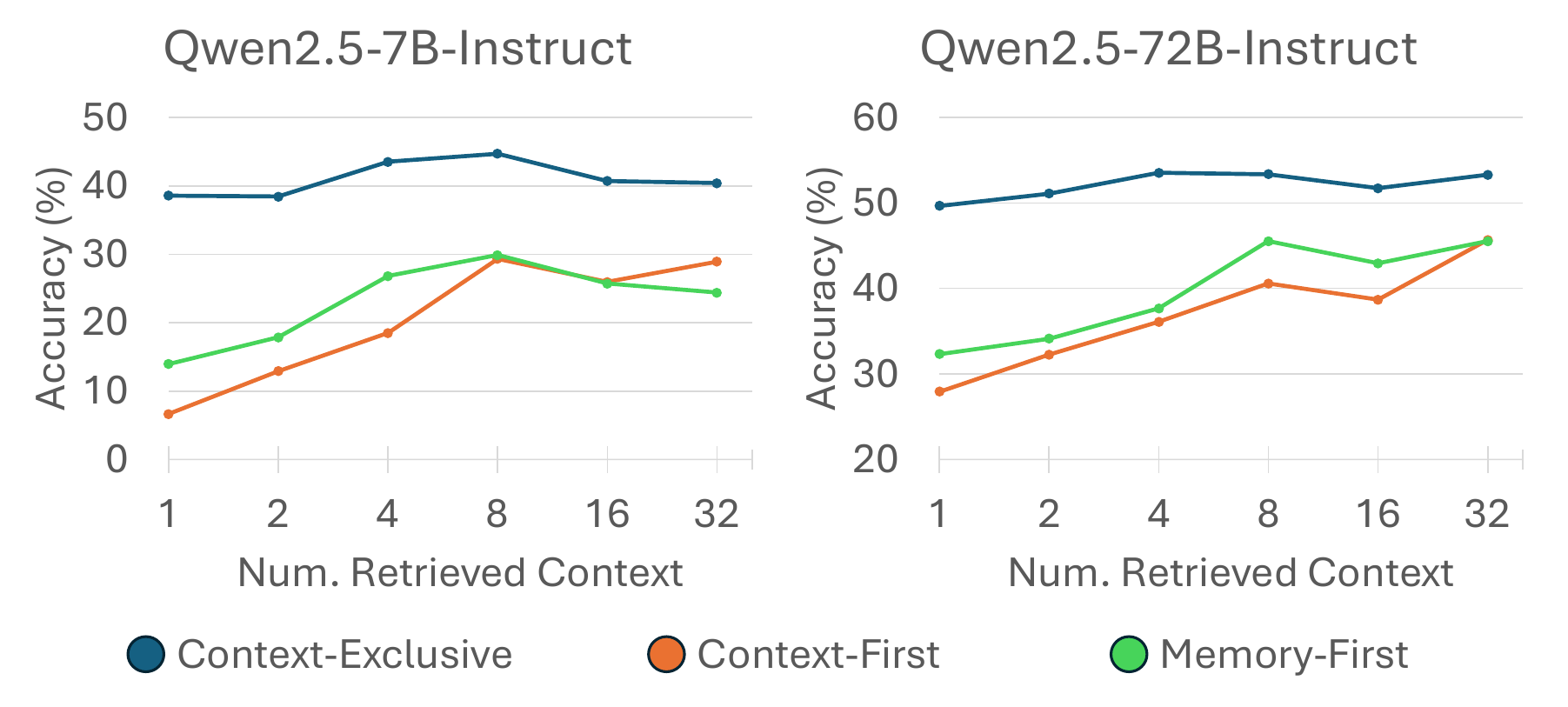}
\caption{Case-Level Accuracy curve of Qwen2.5 on HotpotQA.}
\label{fig:abc_hotpotqa_qwen}
\end{subfigure}

\begin{subfigure}{\columnwidth}
\centering
\includegraphics[width=\columnwidth]{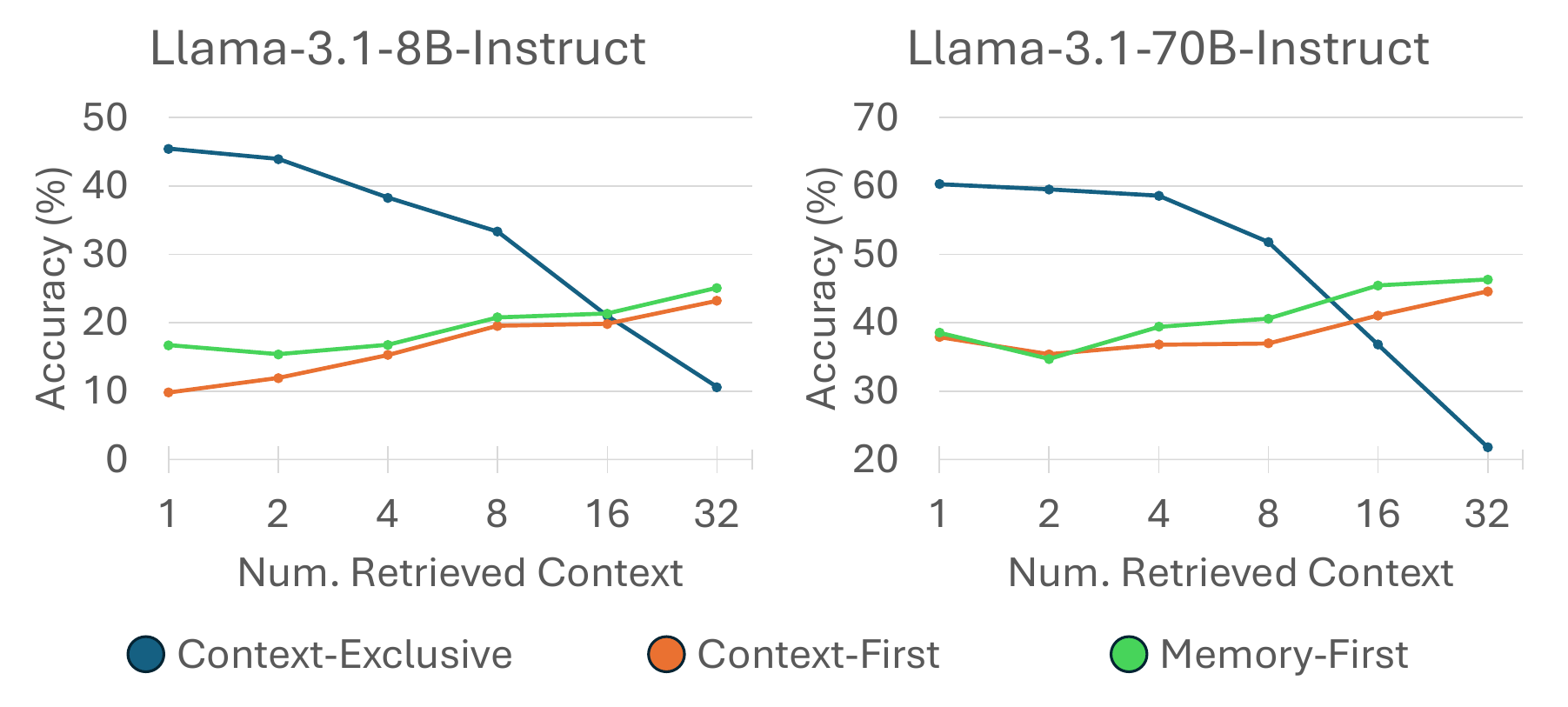}
\caption{Case-Level Accuracy curve of Llama-3.1 on HotpotQA.}
\label{fig:abc_hotpotqa_llama}
\end{subfigure}
\vspace{-0.2in}
\caption{Case-Level Accuracy curve of Qwen2.5 and Llama-3.1 on HotpotQA.}
\vspace{-0.28in}
\label{fig:setting_abc}
\end{figure}

To further analyze the behavior of LMs on each user need, we measure the curve of \emph{Case-Level Accuracy} versus number of retrieved context on HotpotQA, as shown in Figure \ref{fig:setting_abc}. We defer other two datasets to Figure \ref{fig:setting_abc_total} in the Appendix \ref{sec:accuracy_curve_of_all_datasets}.

\paragraph{Restricting memory usage improves real-world performance.} We find that the model's accuracy increased from \emph{Context or Memory-First} to \emph{Context-Exclusive} case, meaning that limiting the usage of internal memory improves the lower limit of general performance, possibly because \emph{Context-Exclusive} strategy forces strict reliance on retrieved evidence and prevents hallucinations. This trend is particularly evident in Qwen2.5 models on HotpotQA dataset that maintain at least 7.7\% increase in accuracy. However, as the number of context increases, the  performance gap gradually shrinks and may even be inverted on Llama-3.1 models where \emph{Context-Exclusive} accuracy drops by up to 12.5\% when the number of retrieved context increases to 32. 

\begin{figure}[htbp!]
\centering
\includegraphics[width=\columnwidth]{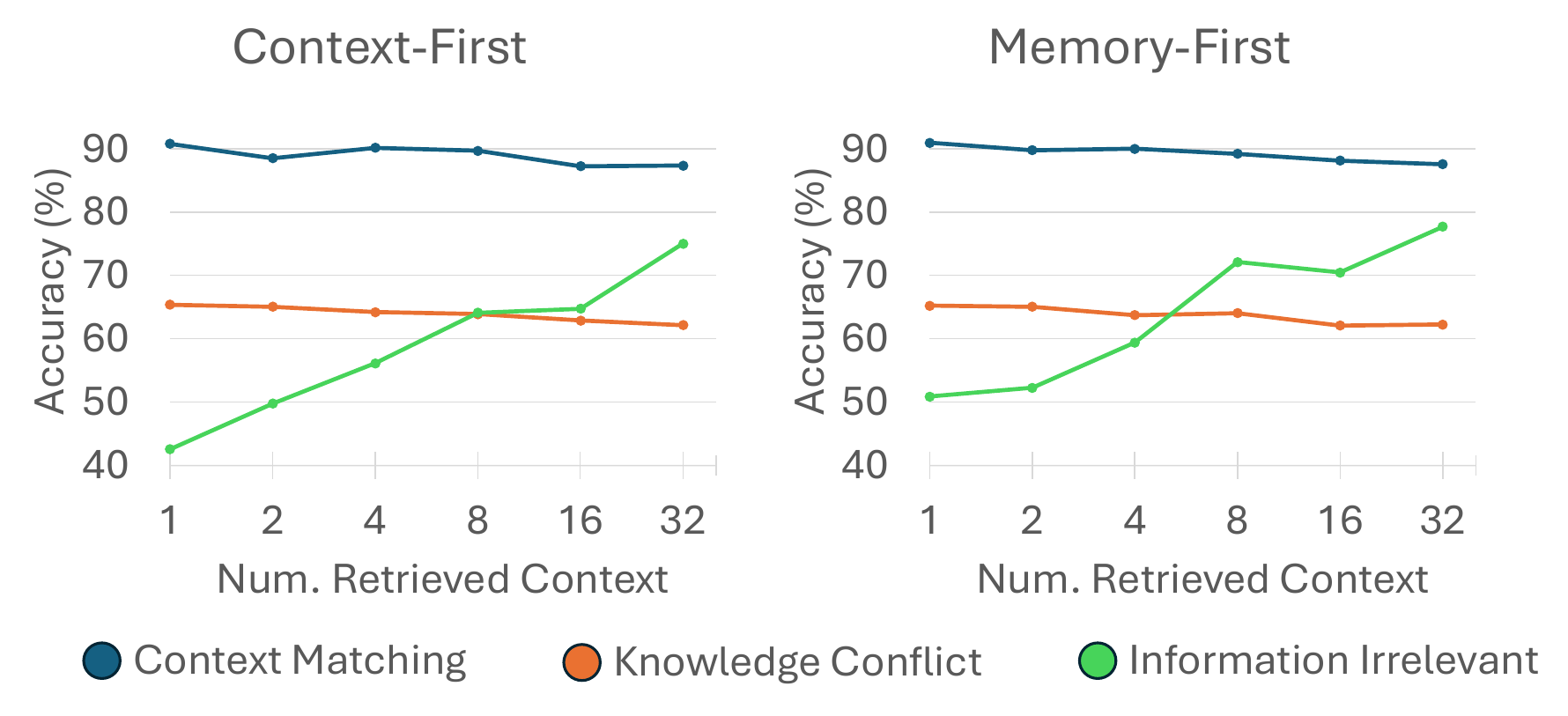}
\caption{Accuracy curve of Qwen2.5-72B-Instruct on HotpotQA dataset under all context settings with \emph{Context-First} and \emph{Memory-First}.}
\vspace{-0.2in}
\label{fig:case23_hotpotqa_qwen72b}
\end{figure}

\paragraph{Models Tend to Be Lazy with More Context.} To investigate the counterintuitive pattern in which the accuracy of \emph{Context or Memory-First} cases increases as the number of retrieved contexts grows across all models, we analyze the impact of different context settings in both cases, as shown in Figure \ref{fig:case23_hotpotqa_qwen72b}.
Interestingly, the \emph{Information Irrelevant} setting appears to contribute to this upward trend. By randomly sampling 100 cases across different retrieval context lengths, we observe that models are easily influenced by irrelevant information, often generating responses such as “no,” “none,” or “0.” However, as more context is retrieved, models exhibit emergent Chain-of-Thought reasoning capabilities. This phenomenon may stem from a form of "lazy" behavior, where models, instead of actively identifying the correct context, increasingly rely on their own memory as the context length grows. We defer the case study example into Appendix \ref{sec:case_study_of_model_lazy}.



\begin{figure}[t!]
\centering

\begin{subfigure}{\columnwidth}
\centering
\includegraphics[width=\columnwidth]{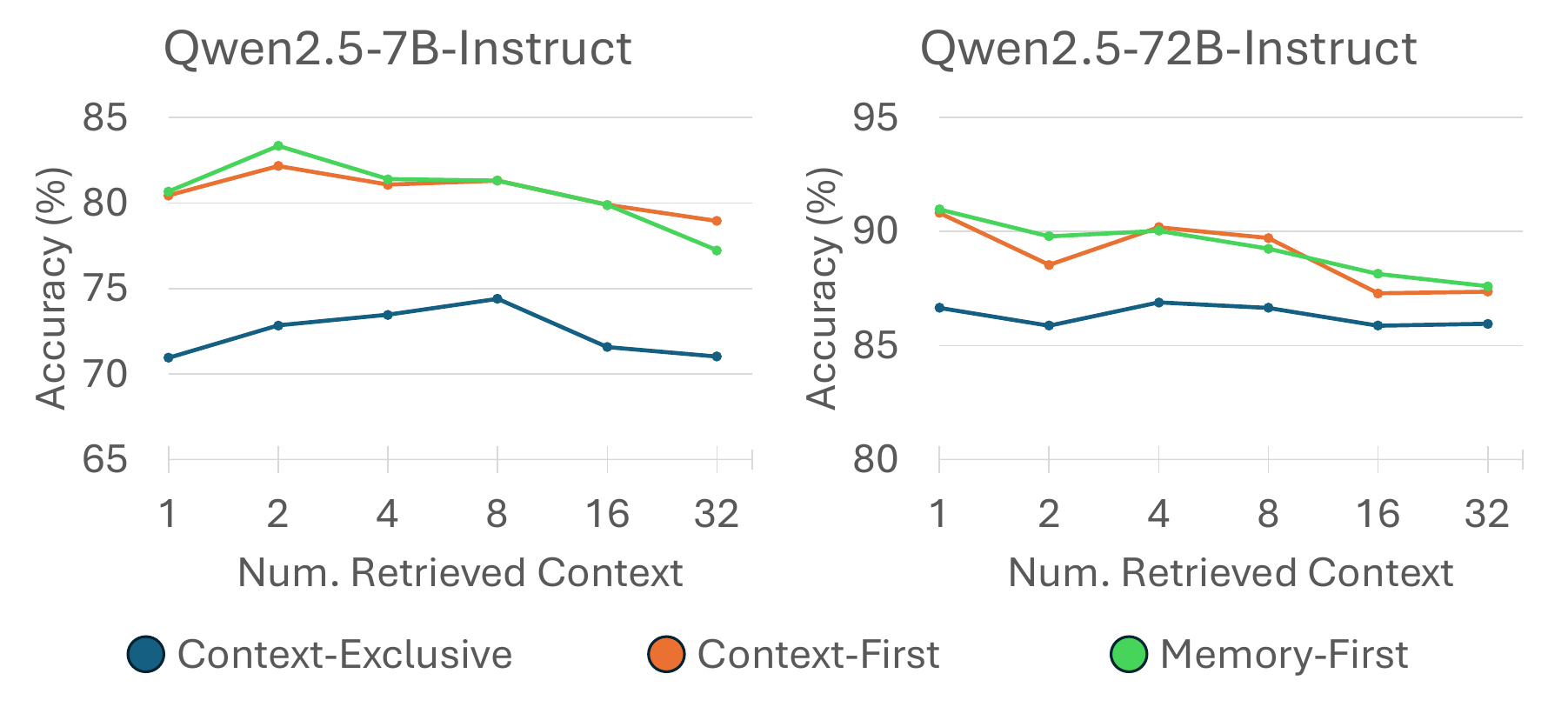}
\caption{Setting-Specific Accuracy curve of Qwen2.5 models on HotpotQA dataset with context matching setting.}
\label{fig:a_disentqa_qwen}
\end{subfigure}

\begin{subfigure}{\columnwidth}
\centering
\includegraphics[width=\columnwidth]{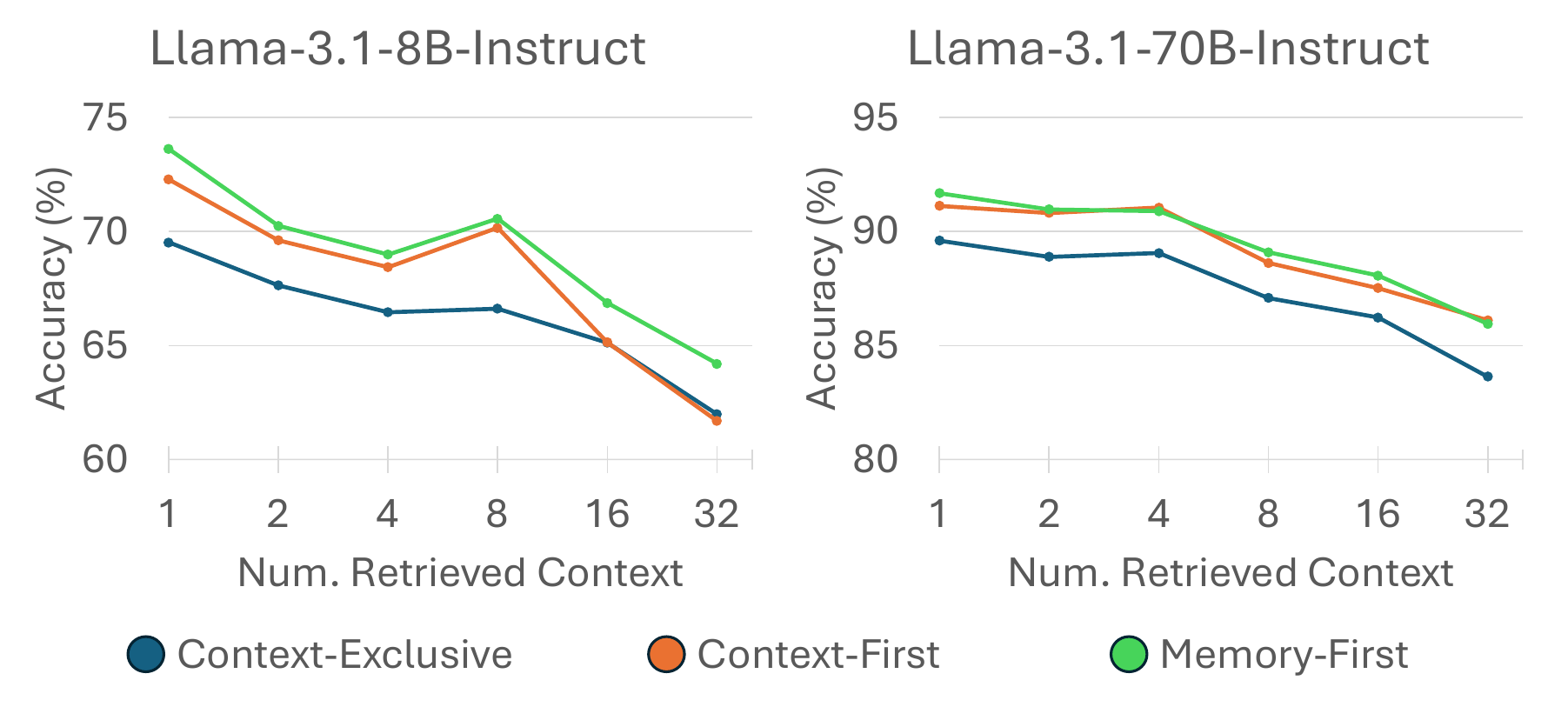}
\caption{Setting-Specific Accuracy curve of Llama-3.1 models on HotpotQA dataset with context matching setting.}
\label{fig:a_disentqa_llama}
\end{subfigure}

\caption{Setting-Specific Accuracy curve of Qwen2.5 and Llama-3.1 models on HotpotQA dataset with context matching setting. These two model as the representative demonstrate the large and small performance drop from \emph{Context or Memory-First} user need to \emph{Context-Exclusive}.}
\vspace{-0.2in}
\label{fig:setting_a}
\end{figure}

\subsection{Individual Setting Performance}
To provide more detailed analysis on models' behavior on the context setting-level, we measure the \emph{Setting-Specific Accuracy} $Acc_c$ curve for each user need case, categorizing them into two groups: \textbf{Optimal Context}, where the provided context aligns with the model's memory, and \textbf{Challenging Context}, where the context is conflicting or irrelevant.

\subsubsection{Performance on Optimal Context}
\label{sssec:general_performance_ideal}
Under the \emph{Context Matching} setting, where the model receives fully relevant and correct context, we assess its maximum potential performance. This defines an \textbf{optimal performance}, isolating the model’s ability to utilize ideal context without retrieval constraints.

\begin{table}[h!]
\centering
\resizebox{\columnwidth}{!}{
\begin{tabular}{c|cc|cc} 
\toprule
\multirow{2}{*}{\bf Dataset} & \multicolumn{2}{c|}{\bf Llama-3.1-Instruct} & \multicolumn{2}{c}{\bf Qwen2.5-Instruct} \\
\cmidrule(lr){2-3} \cmidrule(lr){4-5}
& \bf 8B (\%) & \bf 70B (\%) & \bf 7B (\%) & \bf 72B (\%) \\
\midrule
URAQ & 52 & 74 & 85 & 97 \\
DisentQA & 70 & 84 & 92 & 98 \\
HotpotQA & 63 & 76 & 84 & 95 \\
\bottomrule
\end{tabular}}
\caption{Percentage of errors that is "I don't know" among the shortest 100 randomly selected samples that under \emph{Context Matching} setting that is \textbf{incorrect} for \emph{Context-Exclusive} user need and \textbf{correct} for \emph{Context or Memory-First}. 
}
\vspace{-0.2in}
\label{tab:a_error_type}
\end{table}

\paragraph{Restricting memory usage limits optimal performance.} Based on the results in Figure \ref{fig:setting_a}, we observe that models’ accuracy declines when internal memory is restricted under the \emph{Context-Exclusive} strategy. This effect is more pronounced in the Qwen2.5 family, where Qwen2.5-7B-Instruct experiences up to a 12.1\% accuracy drop from \emph{Context or Memory-First} to \emph{Context-Exclusive}, whereas the Llama-3.1 family shows only a slight decrease, with Llama-3.1-8B-Instruct losing up to 4.1\%.

\paragraph{LLMs exhibit self-protective conservatism. }To examine the accuracy drop under the \emph{Context-Exclusive} setting, we analyze 100 randomly selected cases with up to four retrieved context segments, where the model provides an incorrect answer under \emph{Context-Exclusive} but a correct one under \emph{Context or Memory-First}. Errors are categorized into two types: (1) the model refuses to answer by stating, "I don't know," and (2) the model generates an incorrect hallucinated response. Table \ref{tab:a_error_type} reports the percentage of refusals.

We observe that models overwhelmingly prefer rejection over hallucination when they struggle to locate relevant context, with refusal rates exceeding 50\% across all models and datasets.  This tendency is particularly strong in the Qwen2.5 family, where the 7B and 72B models reject answers in over 85\% of cases, with Qwen2.5-72B-Instruct reaching a 98\% rejection rate on DisentQA. Similarly, the Llama-3.1 models exhibit high rejection rates, ranging from 70\% to 84\% on DisentQA. This conservative behavior may stem from its training objectives or alignment strategies prioritizing answer correctness over speculative responses.


\begin{figure}[h!]
\centering

\begin{subfigure}{\columnwidth}
\centering
\includegraphics[width=\columnwidth]{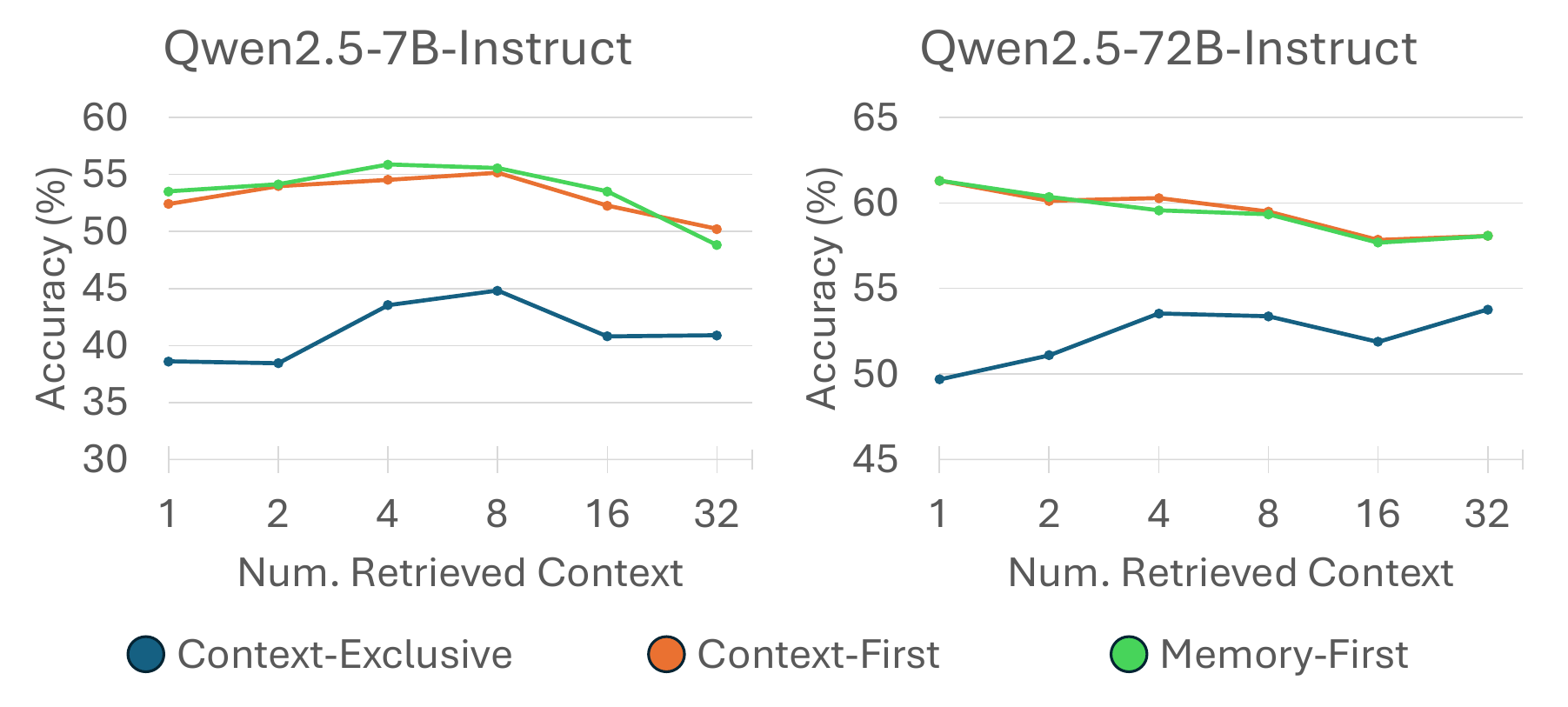}
\caption{Setting-Specific Accuracy curve of Qwen2.5 model family on HotpotQA dataset with knowledge conflict.}
\label{fig:ab_synthetic_qwen}
\end{subfigure}

\begin{subfigure}{\columnwidth}
\centering
\includegraphics[width=\columnwidth]{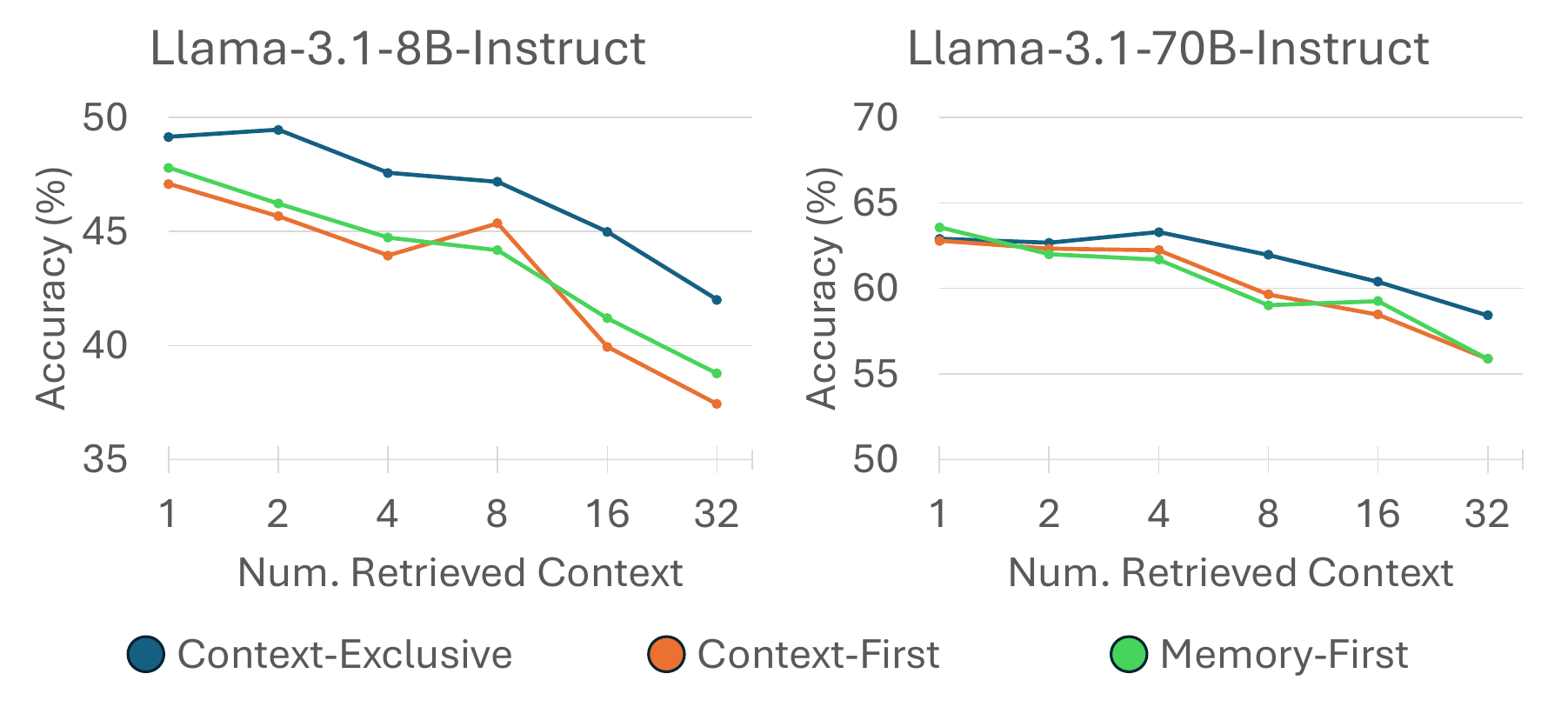}
\caption{Setting-Specific Accuracy curve of Llama-3.1 model family on HotpotQA dataset with knowledge conflict.}
\label{fig:ab_synthetic_llama}
\end{subfigure}

\caption{Setting-Specific Accuracy curve of Qwen2.5 and Llama-3.1 model family on HotpotQA dataset with knowledge conflict. While two models have similar accuracy on \emph{Context or Memory-First} case, Llama models has lower accuracy on \emph{Memory-Exclusive} compared with \emph{Context or Memory-First} and Qwen models has higher accuracy.}
\vspace{-0.2in}
\label{fig:setting_ab}
\end{figure}

\subsubsection{Performance with Challenging Context}
\begin{figure}[htbp!]
\centering

\begin{subfigure}{\columnwidth}
\centering
\includegraphics[width=\columnwidth]{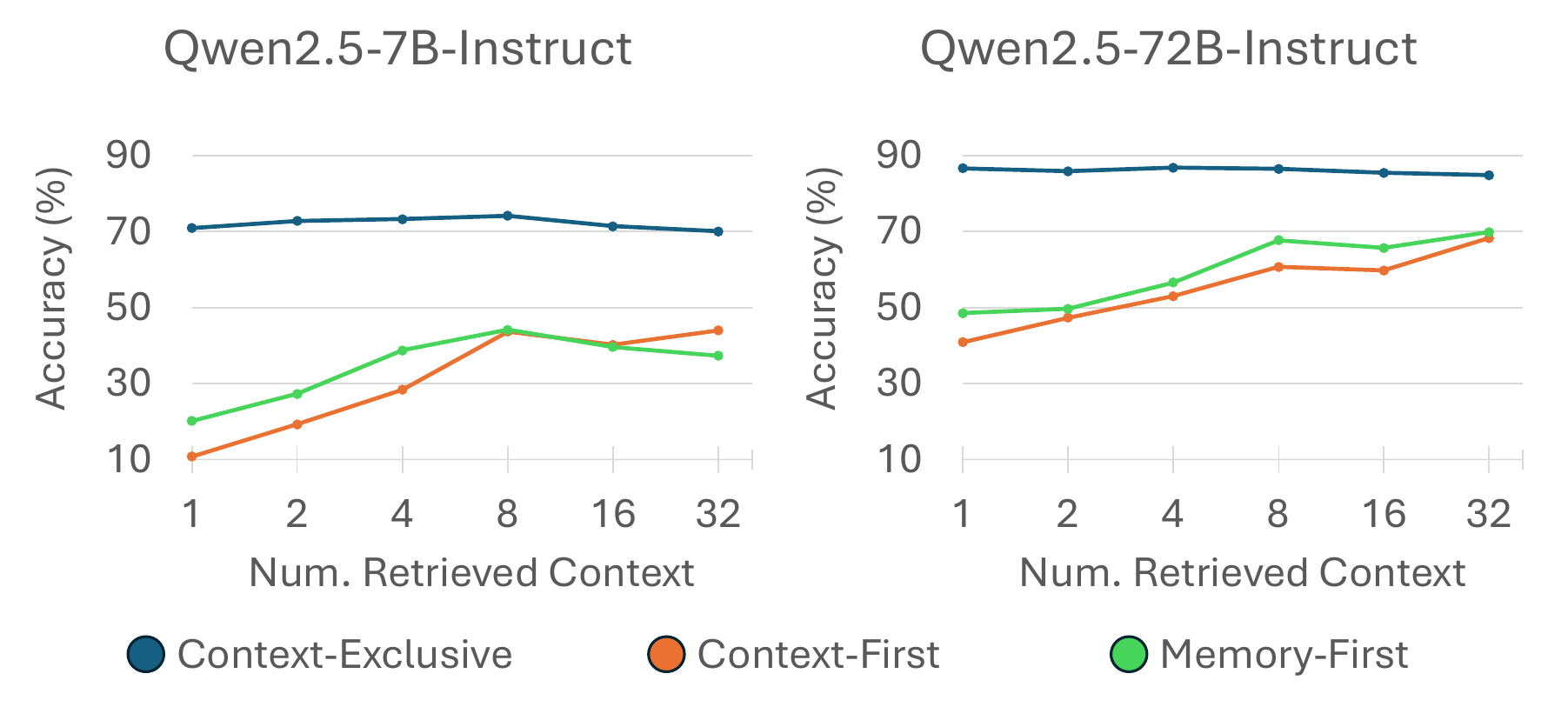}
\caption{Setting-Specific Accuracy curve of Qwen2.5 model family on HotpotQA dataset with irrelevant context.}
\label{fig:ac_hotpotqa_qwen}
\end{subfigure}

\begin{subfigure}{\columnwidth}
\centering
\includegraphics[width=\columnwidth]{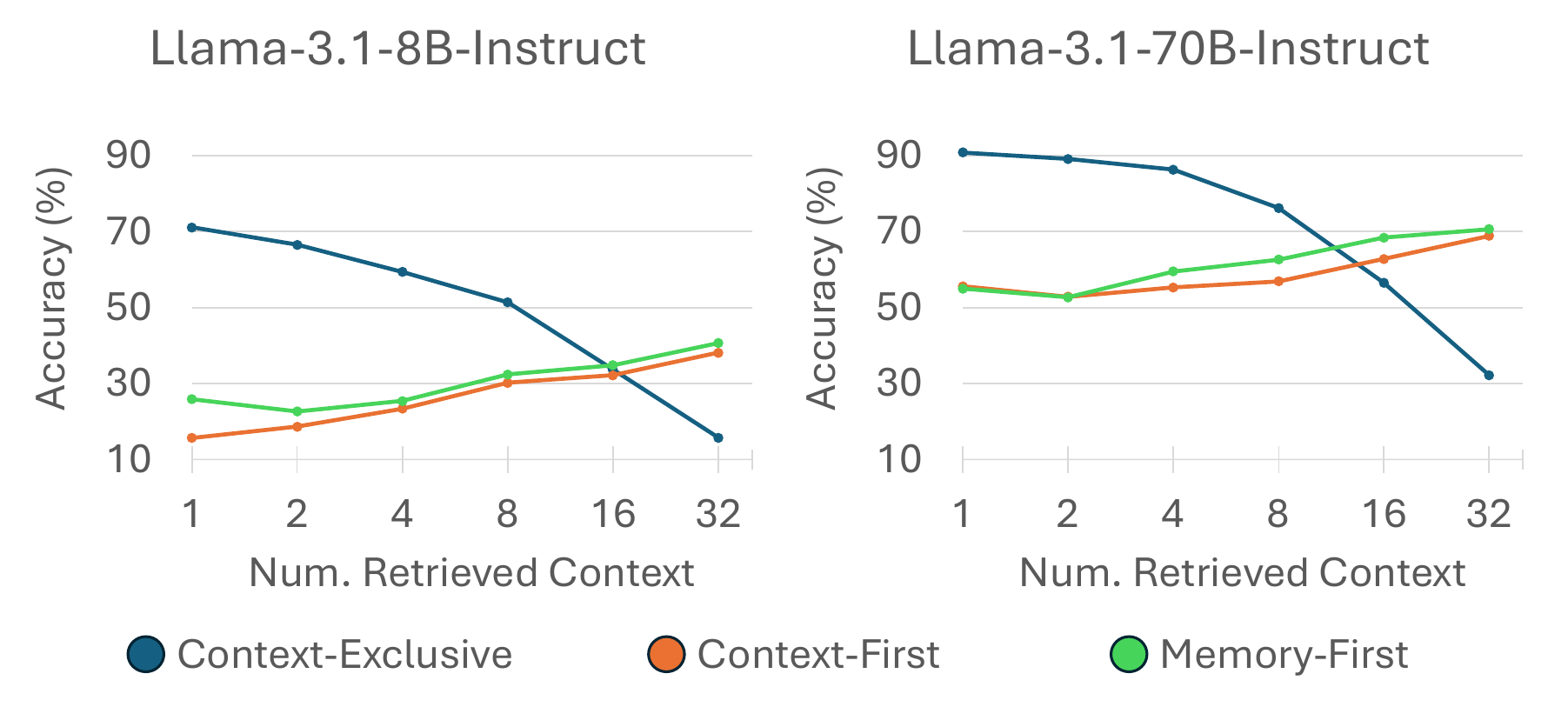}
\caption{Setting-Specific Accuracy curve of Llama-3.1 model family on HotpotQA dataset with irrelevant context.}
\label{fig:ac_hotpotqa_llama}
\end{subfigure}

\caption{Setting-Specific Accuracy curve of Qwen2.5 and Llama-3.1 model family on HotpotQA dataset with irrelevant context.}
\vspace{-0.2in}
\label{fig:setting_ac}
\end{figure}

For performance under \emph{Knowledge Conflict} or \emph{Irrelevant Context}, we realize that evaluating only the performance of single context setting in isolation can introduce bias and skewed interpretations due to LMs preference on using memory than context or vise versa (\citealp{longpre-etal-2021-entity}; \citealp{jin2024tugofwarknowledgeexploringresolving}), resulting performing perfectly in one setting but failed in other. For example, succeeding in \emph{Irrelevant Context} but failing in \emph{Matching Context} may suggest that the model is prone always relying on memory without actually complying with the instructions to use retrieved context. Therefore, we measure the \emph{Setting-Specific Accuracy} $Acc_c$ for Challenging Context in a way that the same question need to be also answered correctly in \emph{Context Matching} settings, ensuring the robustness of evaluation. Such measuring method is applied to all experiments in this section shown in Figure \ref{fig:setting_ab} and \ref{fig:setting_ac}.


\paragraph{Model family dominates behavioral difference.} Model families still exhibit distinct behavioral patterns: When knowledge conflict exists as Figure \ref{fig:setting_ab}, Llama3.1 models show degradation of performance from \emph{Context-First} and \emph{Memory-First} to \emph{Context-Exclusive} case for up to 10.2\% accuracy, while Qwen2.5 models demonstrate the opposite trend with an increase close to 20\%. This behavior suggests fundamental differences in knowledge reliance—Llama3.1 appears more context-dependent, struggling to effectively integrate memory, whereas Qwen2.5 leverages its parametric knowledge more effectively when permitted. Such difference also appears in the as Figure \ref{fig:setting_ac} with \emph{Information Irrelevant} setting, Llama models exhibit significant decreasing accuracy on \emph{Context-Exclusive} strategy with increasing context length for up to 60.1\%, whereas Qwen exhibit almost no loss in performance, for the same reason as discussed in Section \ref{subsec:general_performance_nonideal}.

\section{Conclusion}

We introduce an evaluation framework for RALMs that systematically assesses performance across diverse user needs and context settings. By decomposing user instructions into three generic user need cases (Context-Exclusive, Context-First, Memory-First) and three context settings (Context Matching, Knowledge Conflict, Information Irrelevant), our framework provides comprehensive insights into model capabilities and limitations. Our analysis covers overall user requirements, case-level evaluations, and the impact of varying context contents across different context lengths. The findings highlight the need for user-centric evaluations and architectural innovations to enhance RAG system reliability and real-world applicability.


\section{Limitations}
While our study provides a structured evaluation framework for Retrieval-Augmented Language Models (RALMs) under diverse user needs and retrieval conditions, several limitations remain. Our experiments rely on three datasets: HotpotQA, DisentQA, and the synthetic URAQ dataset. While these datasets cover various knowledge retrieval challenges, they may not fully capture the diversity of real-world retrieval scenarios, particularly in highly specialized domains such as medical or legal applications. Additionally, the synthetic URAQ dataset, although designed to control retrieval complexity, may not generalize perfectly to naturally occurring retrieval conflicts found in real-world settings. In addition, our results are based on evaluations of two model families, Llama-3.1 and Qwen-2.5, across different sizes. While these models are representative of current state-of-the-art retrieval-augmented systems, our conclusions may not generalize to other architectures, such as retrieval-heavy fine-tuned transformers or proprietary models with distinct retrieval and reasoning mechanisms. Future work should extend this analysis to a broader range of models.

\section{Ethics Statement}
Our framework is designed to assess how well RALMs adhere to different user instructions, reflecting real-world applications where users may have distinct expectations regarding knowledge usage. However, models may still exhibit disparities in their ability to satisfy certain user needs, especially in adversarial retrieval settings. We recommend further research on mitigating disparities and enhancing fairness in retrieval-augmented systems. The datasets used in our experiments include HotpotQA, DisentQA, and the newly introduced synthetic URAQ dataset. While these datasets contain diverse question-answer pairs, we acknowledge that biases may be present in both retrieved and internally generated content. We have taken measures to minimize biases by curating synthetic data with balanced question difficulty and by evaluating model performance under varying retrieval conditions. However, residual biases in training corpora or retrieval mechanisms may influence the observed model behavior. One of our primary motivations is to analyze how models handle conflicting or irrelevant retrieved information. While our evaluation reveals scenarios where models fail to distinguish misinformation or exhibit hallucination tendencies, our work does not actively promote the generation or dissemination of false information. Instead, we highlight the need for more robust mechanisms to ensure factual consistency, particularly in knowledge-conflict scenarios. By conducting this study, we aim to advance the ethical design of retrieval-augmented models while encouraging further research on mitigating biases, improving factual robustness, and ensuring alignment with diverse user needs.

\section*{Acknowledgments}
We acknowledge the use of the GPT-4o language model provided by OpenAI in the final stages of manuscript preparation. This tool was employed exclusively for identifying and correcting typographical and grammatical errors, ensuring clarity and precision in the written presentation. Its use was strictly limited to linguistic refinement and did not impact the study’s conceptual framework, research methodology, data analysis, or conclusions. All intellectual contributions and substantive content remain those of the authors.
\bibliography{custom}

\begin{thebibliography}{34}
\providecommand{\natexlab}[1]{#1}

\bibitem[{Ashok and Poczos(2024)}]{ashok2024controllabletextgenerationinstructiontuning}
Dhananjay Ashok and Barnabas Poczos. 2024.
\newblock \href {https://arxiv.org/abs/2405.01490} {Controllable text generation in the instruction-tuning era}.
\newblock \emph{Preprint}, arXiv:2405.01490.

\bibitem[{Borgeaud et~al.(2021)Borgeaud, Mensch, Hoffmann, Cai, Rutherford, Millican, van~den Driessche, Lespiau, Damoc, Clark, de~Las~Casas, Guy, Menick, Ring, Hennigan, Huang, Maggiore, Jones, Cassirer, Brock, Paganini, Irving, Vinyals, Osindero, Simonyan, Rae, Elsen, and Sifre}]{Borgeaud2021ImprovingLM}
Sebastian Borgeaud, Arthur Mensch, Jordan Hoffmann, Trevor Cai, Eliza Rutherford, Katie Millican, George van~den Driessche, Jean-Baptiste Lespiau, Bogdan Damoc, Aidan Clark, Diego de~Las~Casas, Aurelia Guy, Jacob Menick, Roman Ring, T.~W. Hennigan, Saffron Huang, Lorenzo Maggiore, Chris Jones, Albin Cassirer, Andy Brock, Michela Paganini, Geoffrey Irving, Oriol Vinyals, Simon Osindero, Karen Simonyan, Jack~W. Rae, Erich Elsen, and L.~Sifre. 2021.
\newblock \href {https://api.semanticscholar.org/CorpusID:244954723} {Improving language models by retrieving from trillions of tokens}.
\newblock In \emph{International Conference on Machine Learning}.

\bibitem[{Chen et~al.(2024)Chen, Lin, Han, and Sun}]{10.1609/aaai.v38i16.29728}
Jiawei Chen, Hongyu Lin, Xianpei Han, and Le~Sun. 2024.
\newblock \href {https://doi.org/10.1609/aaai.v38i16.29728} {Benchmarking large language models in retrieval-augmented generation}.
\newblock In \emph{Proceedings of the Thirty-Eighth AAAI Conference on Artificial Intelligence and Thirty-Sixth Conference on Innovative Applications of Artificial Intelligence and Fourteenth Symposium on Educational Advances in Artificial Intelligence}, AAAI'24/IAAI'24/EAAI'24. AAAI Press.

\bibitem[{Es et~al.(2023)Es, James, Espinosa-Anke, and Schockaert}]{es2023ragasautomatedevaluationretrieval}
Shahul Es, Jithin James, Luis Espinosa-Anke, and Steven Schockaert. 2023.
\newblock \href {https://arxiv.org/abs/2309.15217} {Ragas: Automated evaluation of retrieval augmented generation}.
\newblock \emph{Preprint}, arXiv:2309.15217.

\bibitem[{Friel et~al.(2024)Friel, Belyi, and Sanyal}]{friel2024ragbenchexplainablebenchmarkretrievalaugmented}
Robert Friel, Masha Belyi, and Atindriyo Sanyal. 2024.
\newblock \href {https://arxiv.org/abs/2407.11005} {Ragbench: Explainable benchmark for retrieval-augmented generation systems}.
\newblock \emph{Preprint}, arXiv:2407.11005.

\bibitem[{Grattafiori et~al.(2024)Grattafiori, Dubey, Jauhri, Pandey, Kadian, Al-Dahle, Letman, Mathur, Schelten, Vaughan, Yang, Fan, Goyal, Hartshorn, Yang, Mitra, Sravankumar, Korenev, Hinsvark, Rao, Zhang, Rodriguez, Gregerson, Spataru, Roziere, Biron, Tang, Chern, Caucheteux, Nayak, Bi, Marra, McConnell, Keller, Touret, Wu, Wong, Ferrer, Nikolaidis, Allonsius, Song, Pintz, Livshits, Wyatt, Esiobu, Choudhary, Mahajan, Garcia-Olano, Perino, Hupkes, Lakomkin, AlBadawy, Lobanova, Dinan, Smith, Radenovic, Guzmán, Zhang, Synnaeve, Lee, Anderson, Thattai, Nail, Mialon, Pang, Cucurell, Nguyen, Korevaar, Xu, Touvron, Zarov, Ibarra, Kloumann, Misra, Evtimov, Zhang, Copet, Lee, Geffert, Vranes, Park, Mahadeokar, Shah, van~der Linde, Billock, Hong, Lee, Fu, Chi, Huang, Liu, Wang, Yu, Bitton, Spisak, Park, Rocca, Johnstun, Saxe, Jia, Alwala, Prasad, Upasani, Plawiak, Li, Heafield, Stone, El-Arini, Iyer, Malik, Chiu, Bhalla, Lakhotia, Rantala-Yeary, van~der Maaten, Chen, Tan, Jenkins, Martin, Madaan, Malo, Blecher,
  Landzaat, de~Oliveira, Muzzi, Pasupuleti, Singh, Paluri, Kardas, Tsimpoukelli, Oldham, Rita, Pavlova, Kambadur, Lewis, Si, Singh, Hassan, Goyal, Torabi, Bashlykov, Bogoychev, Chatterji, Zhang, Duchenne, Çelebi, Alrassy, Zhang, Li, Vasic, Weng, Bhargava, Dubal, Krishnan, Koura, Xu, He, Dong, Srinivasan, Ganapathy, Calderer, Cabral, Stojnic, Raileanu, Maheswari, Girdhar, Patel, Sauvestre, Polidoro, Sumbaly, Taylor, Silva, Hou, Wang, Hosseini, Chennabasappa, Singh, Bell, Kim, Edunov, Nie, Narang, Raparthy, Shen, Wan, Bhosale, Zhang, Vandenhende, Batra, Whitman, Sootla, Collot, Gururangan, Borodinsky, Herman, Fowler, Sheasha, Georgiou, Scialom, Speckbacher, Mihaylov, Xiao, Karn, Goswami, Gupta, Ramanathan, Kerkez, Gonguet, Do, Vogeti, Albiero, Petrovic, Chu, Xiong, Fu, Meers, Martinet, Wang, Wang, Tan, Xia, Xie, Jia, Wang, Goldschlag, Gaur, Babaei, Wen, Song, Zhang, Li, Mao, Coudert, Yan, Chen, Papakipos, Singh, Srivastava, Jain, Kelsey, Shajnfeld, Gangidi, Victoria, Goldstand, Menon, Sharma, Boesenberg,
  Baevski, Feinstein, Kallet, Sangani, Teo, Yunus, Lupu, Alvarado, Caples, Gu, Ho, Poulton, Ryan, Ramchandani, Dong, Franco, Goyal, Saraf, Chowdhury, Gabriel, Bharambe, Eisenman, Yazdan, James, Maurer, Leonhardi, Huang, Loyd, Paola, Paranjape, Liu, Wu, Ni, Hancock, Wasti, Spence, Stojkovic, Gamido, Montalvo, Parker, Burton, Mejia, Liu, Wang, Kim, Zhou, Hu, Chu, Cai, Tindal, Feichtenhofer, Gao, Civin, Beaty, Kreymer, Li, Adkins, Xu, Testuggine, David, Parikh, Liskovich, Foss, Wang, Le, Holland, Dowling, Jamil, Montgomery, Presani, Hahn, Wood, Le, Brinkman, Arcaute, Dunbar, Smothers, Sun, Kreuk, Tian, Kokkinos, Ozgenel, Caggioni, Kanayet, Seide, Florez, Schwarz, Badeer, Swee, Halpern, Herman, Sizov, Guangyi, Zhang, Lakshminarayanan, Inan, Shojanazeri, Zou, Wang, Zha, Habeeb, Rudolph, Suk, Aspegren, Goldman, Zhan, Damlaj, Molybog, Tufanov, Leontiadis, Veliche, Gat, Weissman, Geboski, Kohli, Lam, Asher, Gaya, Marcus, Tang, Chan, Zhen, Reizenstein, Teboul, Zhong, Jin, Yang, Cummings, Carvill, Shepard, McPhie,
  Torres, Ginsburg, Wang, Wu, U, Saxena, Khandelwal, Zand, Matosich, Veeraraghavan, Michelena, Li, Jagadeesh, Huang, Chawla, Huang, Chen, Garg, A, Silva, Bell, Zhang, Guo, Yu, Moshkovich, Wehrstedt, Khabsa, Avalani, Bhatt, Mankus, Hasson, Lennie, Reso, Groshev, Naumov, Lathi, Keneally, Liu, Seltzer, Valko, Restrepo, Patel, Vyatskov, Samvelyan, Clark, Macey, Wang, Hermoso, Metanat, Rastegari, Bansal, Santhanam, Parks, White, Bawa, Singhal, Egebo, Usunier, Mehta, Laptev, Dong, Cheng, Chernoguz, Hart, Salpekar, Kalinli, Kent, Parekh, Saab, Balaji, Rittner, Bontrager, Roux, Dollar, Zvyagina, Ratanchandani, Yuvraj, Liang, Alao, Rodriguez, Ayub, Murthy, Nayani, Mitra, Parthasarathy, Li, Hogan, Battey, Wang, Howes, Rinott, Mehta, Siby, Bondu, Datta, Chugh, Hunt, Dhillon, Sidorov, Pan, Mahajan, Verma, Yamamoto, Ramaswamy, Lindsay, Lindsay, Feng, Lin, Zha, Patil, Shankar, Zhang, Zhang, Wang, Agarwal, Sajuyigbe, Chintala, Max, Chen, Kehoe, Satterfield, Govindaprasad, Gupta, Deng, Cho, Virk, Subramanian, Choudhury,
  Goldman, Remez, Glaser, Best, Koehler, Robinson, Li, Zhang, Matthews, Chou, Shaked, Vontimitta, Ajayi, Montanez, Mohan, Kumar, Mangla, Ionescu, Poenaru, Mihailescu, Ivanov, Li, Wang, Jiang, Bouaziz, Constable, Tang, Wu, Wang, Wu, Gao, Kleinman, Chen, Hu, Jia, Qi, Li, Zhang, Zhang, Adi, Nam, Yu, Wang, Zhao, Hao, Qian, Li, He, Rait, DeVito, Rosnbrick, Wen, Yang, Zhao, and Ma}]{grattafiori2024llama3herdmodels}
Aaron Grattafiori, Abhimanyu Dubey, Abhinav Jauhri, Abhinav Pandey, Abhishek Kadian, Ahmad Al-Dahle, Aiesha Letman, Akhil Mathur, Alan Schelten, Alex Vaughan, Amy Yang, Angela Fan, Anirudh Goyal, Anthony Hartshorn, Aobo Yang, Archi Mitra, Archie Sravankumar, Artem Korenev, Arthur Hinsvark, Arun Rao, Aston Zhang, Aurelien Rodriguez, Austen Gregerson, Ava Spataru, Baptiste Roziere, Bethany Biron, Binh Tang, Bobbie Chern, Charlotte Caucheteux, Chaya Nayak, Chloe Bi, Chris Marra, Chris McConnell, Christian Keller, Christophe Touret, Chunyang Wu, Corinne Wong, Cristian~Canton Ferrer, Cyrus Nikolaidis, Damien Allonsius, Daniel Song, Danielle Pintz, Danny Livshits, Danny Wyatt, David Esiobu, Dhruv Choudhary, Dhruv Mahajan, Diego Garcia-Olano, Diego Perino, Dieuwke Hupkes, Egor Lakomkin, Ehab AlBadawy, Elina Lobanova, Emily Dinan, Eric~Michael Smith, Filip Radenovic, Francisco Guzmán, Frank Zhang, Gabriel Synnaeve, Gabrielle Lee, Georgia~Lewis Anderson, Govind Thattai, Graeme Nail, Gregoire Mialon, Guan Pang,
  Guillem Cucurell, Hailey Nguyen, Hannah Korevaar, Hu~Xu, Hugo Touvron, Iliyan Zarov, Imanol~Arrieta Ibarra, Isabel Kloumann, Ishan Misra, Ivan Evtimov, Jack Zhang, Jade Copet, Jaewon Lee, Jan Geffert, Jana Vranes, Jason Park, Jay Mahadeokar, Jeet Shah, Jelmer van~der Linde, Jennifer Billock, Jenny Hong, Jenya Lee, Jeremy Fu, Jianfeng Chi, Jianyu Huang, Jiawen Liu, Jie Wang, Jiecao Yu, Joanna Bitton, Joe Spisak, Jongsoo Park, Joseph Rocca, Joshua Johnstun, Joshua Saxe, Junteng Jia, Kalyan~Vasuden Alwala, Karthik Prasad, Kartikeya Upasani, Kate Plawiak, Ke~Li, Kenneth Heafield, Kevin Stone, Khalid El-Arini, Krithika Iyer, Kshitiz Malik, Kuenley Chiu, Kunal Bhalla, Kushal Lakhotia, Lauren Rantala-Yeary, Laurens van~der Maaten, Lawrence Chen, Liang Tan, Liz Jenkins, Louis Martin, Lovish Madaan, Lubo Malo, Lukas Blecher, Lukas Landzaat, Luke de~Oliveira, Madeline Muzzi, Mahesh Pasupuleti, Mannat Singh, Manohar Paluri, Marcin Kardas, Maria Tsimpoukelli, Mathew Oldham, Mathieu Rita, Maya Pavlova, Melanie Kambadur,
  Mike Lewis, Min Si, Mitesh~Kumar Singh, Mona Hassan, Naman Goyal, Narjes Torabi, Nikolay Bashlykov, Nikolay Bogoychev, Niladri Chatterji, Ning Zhang, Olivier Duchenne, Onur Çelebi, Patrick Alrassy, Pengchuan Zhang, Pengwei Li, Petar Vasic, Peter Weng, Prajjwal Bhargava, Pratik Dubal, Praveen Krishnan, Punit~Singh Koura, Puxin Xu, Qing He, Qingxiao Dong, Ragavan Srinivasan, Raj Ganapathy, Ramon Calderer, Ricardo~Silveira Cabral, Robert Stojnic, Roberta Raileanu, Rohan Maheswari, Rohit Girdhar, Rohit Patel, Romain Sauvestre, Ronnie Polidoro, Roshan Sumbaly, Ross Taylor, Ruan Silva, Rui Hou, Rui Wang, Saghar Hosseini, Sahana Chennabasappa, Sanjay Singh, Sean Bell, Seohyun~Sonia Kim, Sergey Edunov, Shaoliang Nie, Sharan Narang, Sharath Raparthy, Sheng Shen, Shengye Wan, Shruti Bhosale, Shun Zhang, Simon Vandenhende, Soumya Batra, Spencer Whitman, Sten Sootla, Stephane Collot, Suchin Gururangan, Sydney Borodinsky, Tamar Herman, Tara Fowler, Tarek Sheasha, Thomas Georgiou, Thomas Scialom, Tobias Speckbacher,
  Todor Mihaylov, Tong Xiao, Ujjwal Karn, Vedanuj Goswami, Vibhor Gupta, Vignesh Ramanathan, Viktor Kerkez, Vincent Gonguet, Virginie Do, Vish Vogeti, Vítor Albiero, Vladan Petrovic, Weiwei Chu, Wenhan Xiong, Wenyin Fu, Whitney Meers, Xavier Martinet, Xiaodong Wang, Xiaofang Wang, Xiaoqing~Ellen Tan, Xide Xia, Xinfeng Xie, Xuchao Jia, Xuewei Wang, Yaelle Goldschlag, Yashesh Gaur, Yasmine Babaei, Yi~Wen, Yiwen Song, Yuchen Zhang, Yue Li, Yuning Mao, Zacharie~Delpierre Coudert, Zheng Yan, Zhengxing Chen, Zoe Papakipos, Aaditya Singh, Aayushi Srivastava, Abha Jain, Adam Kelsey, Adam Shajnfeld, Adithya Gangidi, Adolfo Victoria, Ahuva Goldstand, Ajay Menon, Ajay Sharma, Alex Boesenberg, Alexei Baevski, Allie Feinstein, Amanda Kallet, Amit Sangani, Amos Teo, Anam Yunus, Andrei Lupu, Andres Alvarado, Andrew Caples, Andrew Gu, Andrew Ho, Andrew Poulton, Andrew Ryan, Ankit Ramchandani, Annie Dong, Annie Franco, Anuj Goyal, Aparajita Saraf, Arkabandhu Chowdhury, Ashley Gabriel, Ashwin Bharambe, Assaf Eisenman, Azadeh
  Yazdan, Beau James, Ben Maurer, Benjamin Leonhardi, Bernie Huang, Beth Loyd, Beto~De Paola, Bhargavi Paranjape, Bing Liu, Bo~Wu, Boyu Ni, Braden Hancock, Bram Wasti, Brandon Spence, Brani Stojkovic, Brian Gamido, Britt Montalvo, Carl Parker, Carly Burton, Catalina Mejia, Ce~Liu, Changhan Wang, Changkyu Kim, Chao Zhou, Chester Hu, Ching-Hsiang Chu, Chris Cai, Chris Tindal, Christoph Feichtenhofer, Cynthia Gao, Damon Civin, Dana Beaty, Daniel Kreymer, Daniel Li, David Adkins, David Xu, Davide Testuggine, Delia David, Devi Parikh, Diana Liskovich, Didem Foss, Dingkang Wang, Duc Le, Dustin Holland, Edward Dowling, Eissa Jamil, Elaine Montgomery, Eleonora Presani, Emily Hahn, Emily Wood, Eric-Tuan Le, Erik Brinkman, Esteban Arcaute, Evan Dunbar, Evan Smothers, Fei Sun, Felix Kreuk, Feng Tian, Filippos Kokkinos, Firat Ozgenel, Francesco Caggioni, Frank Kanayet, Frank Seide, Gabriela~Medina Florez, Gabriella Schwarz, Gada Badeer, Georgia Swee, Gil Halpern, Grant Herman, Grigory Sizov, Guangyi, Zhang, Guna
  Lakshminarayanan, Hakan Inan, Hamid Shojanazeri, Han Zou, Hannah Wang, Hanwen Zha, Haroun Habeeb, Harrison Rudolph, Helen Suk, Henry Aspegren, Hunter Goldman, Hongyuan Zhan, Ibrahim Damlaj, Igor Molybog, Igor Tufanov, Ilias Leontiadis, Irina-Elena Veliche, Itai Gat, Jake Weissman, James Geboski, James Kohli, Janice Lam, Japhet Asher, Jean-Baptiste Gaya, Jeff Marcus, Jeff Tang, Jennifer Chan, Jenny Zhen, Jeremy Reizenstein, Jeremy Teboul, Jessica Zhong, Jian Jin, Jingyi Yang, Joe Cummings, Jon Carvill, Jon Shepard, Jonathan McPhie, Jonathan Torres, Josh Ginsburg, Junjie Wang, Kai Wu, Kam~Hou U, Karan Saxena, Kartikay Khandelwal, Katayoun Zand, Kathy Matosich, Kaushik Veeraraghavan, Kelly Michelena, Keqian Li, Kiran Jagadeesh, Kun Huang, Kunal Chawla, Kyle Huang, Lailin Chen, Lakshya Garg, Lavender A, Leandro Silva, Lee Bell, Lei Zhang, Liangpeng Guo, Licheng Yu, Liron Moshkovich, Luca Wehrstedt, Madian Khabsa, Manav Avalani, Manish Bhatt, Martynas Mankus, Matan Hasson, Matthew Lennie, Matthias Reso, Maxim
  Groshev, Maxim Naumov, Maya Lathi, Meghan Keneally, Miao Liu, Michael~L. Seltzer, Michal Valko, Michelle Restrepo, Mihir Patel, Mik Vyatskov, Mikayel Samvelyan, Mike Clark, Mike Macey, Mike Wang, Miquel~Jubert Hermoso, Mo~Metanat, Mohammad Rastegari, Munish Bansal, Nandhini Santhanam, Natascha Parks, Natasha White, Navyata Bawa, Nayan Singhal, Nick Egebo, Nicolas Usunier, Nikhil Mehta, Nikolay~Pavlovich Laptev, Ning Dong, Norman Cheng, Oleg Chernoguz, Olivia Hart, Omkar Salpekar, Ozlem Kalinli, Parkin Kent, Parth Parekh, Paul Saab, Pavan Balaji, Pedro Rittner, Philip Bontrager, Pierre Roux, Piotr Dollar, Polina Zvyagina, Prashant Ratanchandani, Pritish Yuvraj, Qian Liang, Rachad Alao, Rachel Rodriguez, Rafi Ayub, Raghotham Murthy, Raghu Nayani, Rahul Mitra, Rangaprabhu Parthasarathy, Raymond Li, Rebekkah Hogan, Robin Battey, Rocky Wang, Russ Howes, Ruty Rinott, Sachin Mehta, Sachin Siby, Sai~Jayesh Bondu, Samyak Datta, Sara Chugh, Sara Hunt, Sargun Dhillon, Sasha Sidorov, Satadru Pan, Saurabh Mahajan,
  Saurabh Verma, Seiji Yamamoto, Sharadh Ramaswamy, Shaun Lindsay, Shaun Lindsay, Sheng Feng, Shenghao Lin, Shengxin~Cindy Zha, Shishir Patil, Shiva Shankar, Shuqiang Zhang, Shuqiang Zhang, Sinong Wang, Sneha Agarwal, Soji Sajuyigbe, Soumith Chintala, Stephanie Max, Stephen Chen, Steve Kehoe, Steve Satterfield, Sudarshan Govindaprasad, Sumit Gupta, Summer Deng, Sungmin Cho, Sunny Virk, Suraj Subramanian, Sy~Choudhury, Sydney Goldman, Tal Remez, Tamar Glaser, Tamara Best, Thilo Koehler, Thomas Robinson, Tianhe Li, Tianjun Zhang, Tim Matthews, Timothy Chou, Tzook Shaked, Varun Vontimitta, Victoria Ajayi, Victoria Montanez, Vijai Mohan, Vinay~Satish Kumar, Vishal Mangla, Vlad Ionescu, Vlad Poenaru, Vlad~Tiberiu Mihailescu, Vladimir Ivanov, Wei Li, Wenchen Wang, Wenwen Jiang, Wes Bouaziz, Will Constable, Xiaocheng Tang, Xiaojian Wu, Xiaolan Wang, Xilun Wu, Xinbo Gao, Yaniv Kleinman, Yanjun Chen, Ye~Hu, Ye~Jia, Ye~Qi, Yenda Li, Yilin Zhang, Ying Zhang, Yossi Adi, Youngjin Nam, Yu, Wang, Yu~Zhao, Yuchen Hao, Yundi
  Qian, Yunlu Li, Yuzi He, Zach Rait, Zachary DeVito, Zef Rosnbrick, Zhaoduo Wen, Zhenyu Yang, Zhiwei Zhao, and Zhiyu Ma. 2024.
\newblock \href {https://arxiv.org/abs/2407.21783} {The llama 3 herd of models}.
\newblock \emph{Preprint}, arXiv:2407.21783.

\bibitem[{Guu et~al.(2020)Guu, Lee, Tung, Pasupat, and Chang}]{10.5555/3524938.3525306}
Kelvin Guu, Kenton Lee, Zora Tung, Panupong Pasupat, and Ming-Wei Chang. 2020.
\newblock Realm: retrieval-augmented language model pre-training.
\newblock In \emph{Proceedings of the 37th International Conference on Machine Learning}, ICML'20. JMLR.org.

\bibitem[{Han et~al.(2024)Han, Lin, Gurung, Thomas, Chen, Borchers, Gupta, and Koedinger}]{han2024improvingassessmenttutoringpractices}
Zifei~FeiFei Han, Jionghao Lin, Ashish Gurung, Danielle~R. Thomas, Eason Chen, Conrad Borchers, Shivang Gupta, and Kenneth~R. Koedinger. 2024.
\newblock \href {https://arxiv.org/abs/2402.14594} {Improving assessment of tutoring practices using retrieval-augmented generation}.
\newblock \emph{Preprint}, arXiv:2402.14594.

\bibitem[{Islam et~al.(2024)Islam, Rahman, Hossain, Hoque, Joty, and Parvez}]{islam-etal-2024-open}
Shayekh~Bin Islam, Md~Asib Rahman, K~S M~Tozammel Hossain, Enamul Hoque, Shafiq Joty, and Md~Rizwan Parvez. 2024.
\newblock \href {https://doi.org/10.18653/v1/2024.findings-emnlp.831} {Open-{RAG}: Enhanced retrieval augmented reasoning with open-source large language models}.
\newblock In \emph{Findings of the Association for Computational Linguistics: EMNLP 2024}, pages 14231--14244, Miami, Florida, USA. Association for Computational Linguistics.

\bibitem[{Izacard et~al.(2023)Izacard, Lewis, Lomeli, Hosseini, Petroni, Schick, Dwivedi-Yu, Joulin, Riedel, and Grave}]{10.5555/3648699.3648950}
Gautier Izacard, Patrick Lewis, Maria Lomeli, Lucas Hosseini, Fabio Petroni, Timo Schick, Jane Dwivedi-Yu, Armand Joulin, Sebastian Riedel, and Edouard Grave. 2023.
\newblock Atlas: few-shot learning with retrieval augmented language models.
\newblock \emph{J. Mach. Learn. Res.}, 24(1).

\bibitem[{Jin et~al.(2024)Jin, Cao, Chen, Liu, Jiang, Xu, Li, and Zhao}]{jin2024tugofwarknowledgeexploringresolving}
Zhuoran Jin, Pengfei Cao, Yubo Chen, Kang Liu, Xiaojian Jiang, Jiexin Xu, Qiuxia Li, and Jun Zhao. 2024.
\newblock \href {https://arxiv.org/abs/2402.14409} {Tug-of-war between knowledge: Exploring and resolving knowledge conflicts in retrieval-augmented language models}.
\newblock \emph{Preprint}, arXiv:2402.14409.

\bibitem[{Karpukhin et~al.(2020)Karpukhin, Oguz, Min, Lewis, Wu, Edunov, Chen, and Yih}]{karpukhin-etal-2020-dense}
Vladimir Karpukhin, Barlas Oguz, Sewon Min, Patrick Lewis, Ledell Wu, Sergey Edunov, Danqi Chen, and Wen-tau Yih. 2020.
\newblock \href {https://doi.org/10.18653/v1/2020.emnlp-main.550} {Dense passage retrieval for open-domain question answering}.
\newblock In \emph{Proceedings of the 2020 Conference on Empirical Methods in Natural Language Processing (EMNLP)}, pages 6769--6781, Online. Association for Computational Linguistics.

\bibitem[{Laban et~al.(2024)Laban, Fabbri, Xiong, and Wu}]{laban2024summaryhaystackchallengelongcontext}
Philippe Laban, Alexander~R. Fabbri, Caiming Xiong, and Chien-Sheng Wu. 2024.
\newblock \href {https://arxiv.org/abs/2407.01370} {Summary of a haystack: A challenge to long-context llms and rag systems}.
\newblock \emph{Preprint}, arXiv:2407.01370.

\bibitem[{Lewis et~al.(2020)Lewis, Perez, Piktus, Petroni, Karpukhin, Goyal, K\"{u}ttler, Lewis, Yih, Rockt\"{a}schel, Riedel, and Kiela}]{10.5555/3495724.3496517}
Patrick Lewis, Ethan Perez, Aleksandra Piktus, Fabio Petroni, Vladimir Karpukhin, Naman Goyal, Heinrich K\"{u}ttler, Mike Lewis, Wen-tau Yih, Tim Rockt\"{a}schel, Sebastian Riedel, and Douwe Kiela. 2020.
\newblock Retrieval-augmented generation for knowledge-intensive nlp tasks.
\newblock In \emph{Proceedings of the 34th International Conference on Neural Information Processing Systems}, NIPS '20, Red Hook, NY, USA. Curran Associates Inc.

\bibitem[{Li et~al.(2023)Li, Rawat, Zaheer, Wang, Lukasik, Veit, Yu, and Kumar}]{li-etal-2023-large}
Daliang Li, Ankit~Singh Rawat, Manzil Zaheer, Xin Wang, Michal Lukasik, Andreas Veit, Felix Yu, and Sanjiv Kumar. 2023.
\newblock \href {https://doi.org/10.18653/v1/2023.findings-acl.112} {Large language models with controllable working memory}.
\newblock In \emph{Findings of the Association for Computational Linguistics: ACL 2023}, pages 1774--1793, Toronto, Canada. Association for Computational Linguistics.

\bibitem[{Longpre et~al.(2021)Longpre, Perisetla, Chen, Ramesh, DuBois, and Singh}]{longpre-etal-2021-entity}
Shayne Longpre, Kartik Perisetla, Anthony Chen, Nikhil Ramesh, Chris DuBois, and Sameer Singh. 2021.
\newblock \href {https://doi.org/10.18653/v1/2021.emnlp-main.565} {Entity-based knowledge conflicts in question answering}.
\newblock In \emph{Proceedings of the 2021 Conference on Empirical Methods in Natural Language Processing}, pages 7052--7063, Online and Punta Cana, Dominican Republic. Association for Computational Linguistics.

\bibitem[{Miao et~al.(2024)Miao, Thongprayoon, Suppadungsuk, Valencia, and Cheungpasitporn}]{Miao2024IntegratingRG}
Jing Miao, Charat Thongprayoon, Supawadee Suppadungsuk, Oscar A.~Garcia Valencia, and Wisit Cheungpasitporn. 2024.
\newblock \href {https://api.semanticscholar.org/CorpusID:268728365} {Integrating retrieval-augmented generation with large language models in nephrology: Advancing practical applications}.
\newblock \emph{Medicina}, 60.

\bibitem[{Ming et~al.(2025)Ming, Purushwalkam, Pandit, Ke, Nguyen, Xiong, and Joty}]{ming2025faithevallanguagemodelstay}
Yifei Ming, Senthil Purushwalkam, Shrey Pandit, Zixuan Ke, Xuan-Phi Nguyen, Caiming Xiong, and Shafiq Joty. 2025.
\newblock \href {https://arxiv.org/abs/2410.03727} {Faitheval: Can your language model stay faithful to context, even if "the moon is made of marshmallows"}.
\newblock \emph{Preprint}, arXiv:2410.03727.

\bibitem[{Neeman et~al.(2023)Neeman, Aharoni, Honovich, Choshen, Szpektor, and Abend}]{neeman-etal-2023-disentqa}
Ella Neeman, Roee Aharoni, Or~Honovich, Leshem Choshen, Idan Szpektor, and Omri Abend. 2023.
\newblock \href {https://doi.org/10.18653/v1/2023.acl-long.559} {{D}isent{QA}: Disentangling parametric and contextual knowledge with counterfactual question answering}.
\newblock In \emph{Proceedings of the 61st Annual Meeting of the Association for Computational Linguistics (Volume 1: Long Papers)}, pages 10056--10070, Toronto, Canada. Association for Computational Linguistics.

\bibitem[{OpenAI et~al.(2024)OpenAI, :, Hurst, Lerer, Goucher, Perelman, Ramesh, Clark, Ostrow, Welihinda, Hayes, Radford, Mądry, Baker-Whitcomb, Beutel, Borzunov, Carney, Chow, Kirillov, Nichol, Paino, Renzin, Passos, Kirillov, Christakis, Conneau, Kamali, Jabri, Moyer, Tam, Crookes, Tootoochian, Tootoonchian, Kumar, Vallone, Karpathy, Braunstein, Cann, Codispoti, Galu, Kondrich, Tulloch, Mishchenko, Baek, Jiang, Pelisse, Woodford, Gosalia, Dhar, Pantuliano, Nayak, Oliver, Zoph, Ghorbani, Leimberger, Rossen, Sokolowsky, Wang, Zweig, Hoover, Samic, McGrew, Spero, Giertler, Cheng, Lightcap, Walkin, Quinn, Guarraci, Hsu, Kellogg, Eastman, Lugaresi, Wainwright, Bassin, Hudson, Chu, Nelson, Li, Shern, Conger, Barette, Voss, Ding, Lu, Zhang, Beaumont, Hallacy, Koch, Gibson, Kim, Choi, McLeavey, Hesse, Fischer, Winter, Czarnecki, Jarvis, Wei, Koumouzelis, Sherburn, Kappler, Levin, Levy, Carr, Farhi, Mely, Robinson, Sasaki, Jin, Valladares, Tsipras, Li, Nguyen, Findlay, Oiwoh, Wong, Asdar, Proehl, Yang, Antonow,
  Kramer, Peterson, Sigler, Wallace, Brevdo, Mays, Khorasani, Such, Raso, Zhang, von Lohmann, Sulit, Goh, Oden, Salmon, Starace, Brockman, Salman, Bao, Hu, Wong, Wang, Schmidt, Whitney, Jun, Kirchner, de~Oliveira~Pinto, Ren, Chang, Chung, Kivlichan, O'Connell, O'Connell, Osband, Silber, Sohl, Okuyucu, Lan, Kostrikov, Sutskever, Kanitscheider, Gulrajani, Coxon, Menick, Pachocki, Aung, Betker, Crooks, Lennon, Kiros, Leike, Park, Kwon, Phang, Teplitz, Wei, Wolfe, Chen, Harris, Varavva, Lee, Shieh, Lin, Yu, Weng, Tang, Yu, Jang, Candela, Beutler, Landers, Parish, Heidecke, Schulman, Lachman, McKay, Uesato, Ward, Kim, Huizinga, Sitkin, Kraaijeveld, Gross, Kaplan, Snyder, Achiam, Jiao, Lee, Zhuang, Harriman, Fricke, Hayashi, Singhal, Shi, Karthik, Wood, Rimbach, Hsu, Nguyen, Gu-Lemberg, Button, Liu, Howe, Muthukumar, Luther, Ahmad, Kai, Itow, Workman, Pathak, Chen, Jing, Guy, Fedus, Zhou, Mamitsuka, Weng, McCallum, Held, Ouyang, Feuvrier, Zhang, Kondraciuk, Kaiser, Hewitt, Metz, Doshi, Aflak, Simens, Boyd,
  Thompson, Dukhan, Chen, Gray, Hudnall, Zhang, Aljubeh, Litwin, Zeng, Johnson, Shetty, Gupta, Shah, Yatbaz, Yang, Zhong, Glaese, Chen, Janner, Lampe, Petrov, Wu, Wang, Fradin, Pokrass, Castro, de~Castro, Pavlov, Brundage, Wang, Khan, Murati, Bavarian, Lin, Yesildal, Soto, Gimelshein, Cone, Staudacher, Summers, LaFontaine, Chowdhury, Ryder, Stathas, Turley, Tezak, Felix, Kudige, Keskar, Deutsch, Bundick, Puckett, Nachum, Okelola, Boiko, Murk, Jaffe, Watkins, Godement, Campbell-Moore, Chao, McMillan, Belov, Su, Bak, Bakkum, Deng, Dolan, Hoeschele, Welinder, Tillet, Pronin, Tillet, Dhariwal, Yuan, Dias, Lim, Arora, Troll, Lin, Lopes, Puri, Miyara, Leike, Gaubert, Zamani, Wang, Donnelly, Honsby, Smith, Sahai, Ramchandani, Huet, Carmichael, Zellers, Chen, Chen, Nigmatullin, Cheu, Jain, Altman, Schoenholz, Toizer, Miserendino, Agarwal, Culver, Ethersmith, Gray, Grove, Metzger, Hermani, Jain, Zhao, Wu, Jomoto, Wu, Shuaiqi, Xia, Phene, Papay, Narayanan, Coffey, Lee, Hall, Balaji, Broda, Stramer, Xu, Gogineni,
  Christianson, Sanders, Patwardhan, Cunninghman, Degry, Dimson, Raoux, Shadwell, Zheng, Underwood, Markov, Sherbakov, Rubin, Stasi, Kaftan, Heywood, Peterson, Walters, Eloundou, Qi, Moeller, Monaco, Kuo, Fomenko, Chang, Zheng, Zhou, Manassra, Sheu, Zaremba, Patil, Qian, Kim, Cheng, Zhang, He, Zhang, Jin, Dai, and Malkov}]{openai2024gpt4ocard}
OpenAI, :, Aaron Hurst, Adam Lerer, Adam~P. Goucher, Adam Perelman, Aditya Ramesh, Aidan Clark, AJ~Ostrow, Akila Welihinda, Alan Hayes, Alec Radford, Aleksander Mądry, Alex Baker-Whitcomb, Alex Beutel, Alex Borzunov, Alex Carney, Alex Chow, Alex Kirillov, Alex Nichol, Alex Paino, Alex Renzin, Alex~Tachard Passos, Alexander Kirillov, Alexi Christakis, Alexis Conneau, Ali Kamali, Allan Jabri, Allison Moyer, Allison Tam, Amadou Crookes, Amin Tootoochian, Amin Tootoonchian, Ananya Kumar, Andrea Vallone, Andrej Karpathy, Andrew Braunstein, Andrew Cann, Andrew Codispoti, Andrew Galu, Andrew Kondrich, Andrew Tulloch, Andrey Mishchenko, Angela Baek, Angela Jiang, Antoine Pelisse, Antonia Woodford, Anuj Gosalia, Arka Dhar, Ashley Pantuliano, Avi Nayak, Avital Oliver, Barret Zoph, Behrooz Ghorbani, Ben Leimberger, Ben Rossen, Ben Sokolowsky, Ben Wang, Benjamin Zweig, Beth Hoover, Blake Samic, Bob McGrew, Bobby Spero, Bogo Giertler, Bowen Cheng, Brad Lightcap, Brandon Walkin, Brendan Quinn, Brian Guarraci, Brian Hsu,
  Bright Kellogg, Brydon Eastman, Camillo Lugaresi, Carroll Wainwright, Cary Bassin, Cary Hudson, Casey Chu, Chad Nelson, Chak Li, Chan~Jun Shern, Channing Conger, Charlotte Barette, Chelsea Voss, Chen Ding, Cheng Lu, Chong Zhang, Chris Beaumont, Chris Hallacy, Chris Koch, Christian Gibson, Christina Kim, Christine Choi, Christine McLeavey, Christopher Hesse, Claudia Fischer, Clemens Winter, Coley Czarnecki, Colin Jarvis, Colin Wei, Constantin Koumouzelis, Dane Sherburn, Daniel Kappler, Daniel Levin, Daniel Levy, David Carr, David Farhi, David Mely, David Robinson, David Sasaki, Denny Jin, Dev Valladares, Dimitris Tsipras, Doug Li, Duc~Phong Nguyen, Duncan Findlay, Edede Oiwoh, Edmund Wong, Ehsan Asdar, Elizabeth Proehl, Elizabeth Yang, Eric Antonow, Eric Kramer, Eric Peterson, Eric Sigler, Eric Wallace, Eugene Brevdo, Evan Mays, Farzad Khorasani, Felipe~Petroski Such, Filippo Raso, Francis Zhang, Fred von Lohmann, Freddie Sulit, Gabriel Goh, Gene Oden, Geoff Salmon, Giulio Starace, Greg Brockman, Hadi
  Salman, Haiming Bao, Haitang Hu, Hannah Wong, Haoyu Wang, Heather Schmidt, Heather Whitney, Heewoo Jun, Hendrik Kirchner, Henrique~Ponde de~Oliveira~Pinto, Hongyu Ren, Huiwen Chang, Hyung~Won Chung, Ian Kivlichan, Ian O'Connell, Ian O'Connell, Ian Osband, Ian Silber, Ian Sohl, Ibrahim Okuyucu, Ikai Lan, Ilya Kostrikov, Ilya Sutskever, Ingmar Kanitscheider, Ishaan Gulrajani, Jacob Coxon, Jacob Menick, Jakub Pachocki, James Aung, James Betker, James Crooks, James Lennon, Jamie Kiros, Jan Leike, Jane Park, Jason Kwon, Jason Phang, Jason Teplitz, Jason Wei, Jason Wolfe, Jay Chen, Jeff Harris, Jenia Varavva, Jessica~Gan Lee, Jessica Shieh, Ji~Lin, Jiahui Yu, Jiayi Weng, Jie Tang, Jieqi Yu, Joanne Jang, Joaquin~Quinonero Candela, Joe Beutler, Joe Landers, Joel Parish, Johannes Heidecke, John Schulman, Jonathan Lachman, Jonathan McKay, Jonathan Uesato, Jonathan Ward, Jong~Wook Kim, Joost Huizinga, Jordan Sitkin, Jos Kraaijeveld, Josh Gross, Josh Kaplan, Josh Snyder, Joshua Achiam, Joy Jiao, Joyce Lee, Juntang
  Zhuang, Justyn Harriman, Kai Fricke, Kai Hayashi, Karan Singhal, Katy Shi, Kavin Karthik, Kayla Wood, Kendra Rimbach, Kenny Hsu, Kenny Nguyen, Keren Gu-Lemberg, Kevin Button, Kevin Liu, Kiel Howe, Krithika Muthukumar, Kyle Luther, Lama Ahmad, Larry Kai, Lauren Itow, Lauren Workman, Leher Pathak, Leo Chen, Li~Jing, Lia Guy, Liam Fedus, Liang Zhou, Lien Mamitsuka, Lilian Weng, Lindsay McCallum, Lindsey Held, Long Ouyang, Louis Feuvrier, Lu~Zhang, Lukas Kondraciuk, Lukasz Kaiser, Luke Hewitt, Luke Metz, Lyric Doshi, Mada Aflak, Maddie Simens, Madelaine Boyd, Madeleine Thompson, Marat Dukhan, Mark Chen, Mark Gray, Mark Hudnall, Marvin Zhang, Marwan Aljubeh, Mateusz Litwin, Matthew Zeng, Max Johnson, Maya Shetty, Mayank Gupta, Meghan Shah, Mehmet Yatbaz, Meng~Jia Yang, Mengchao Zhong, Mia Glaese, Mianna Chen, Michael Janner, Michael Lampe, Michael Petrov, Michael Wu, Michele Wang, Michelle Fradin, Michelle Pokrass, Miguel Castro, Miguel Oom~Temudo de~Castro, Mikhail Pavlov, Miles Brundage, Miles Wang, Minal
  Khan, Mira Murati, Mo~Bavarian, Molly Lin, Murat Yesildal, Nacho Soto, Natalia Gimelshein, Natalie Cone, Natalie Staudacher, Natalie Summers, Natan LaFontaine, Neil Chowdhury, Nick Ryder, Nick Stathas, Nick Turley, Nik Tezak, Niko Felix, Nithanth Kudige, Nitish Keskar, Noah Deutsch, Noel Bundick, Nora Puckett, Ofir Nachum, Ola Okelola, Oleg Boiko, Oleg Murk, Oliver Jaffe, Olivia Watkins, Olivier Godement, Owen Campbell-Moore, Patrick Chao, Paul McMillan, Pavel Belov, Peng Su, Peter Bak, Peter Bakkum, Peter Deng, Peter Dolan, Peter Hoeschele, Peter Welinder, Phil Tillet, Philip Pronin, Philippe Tillet, Prafulla Dhariwal, Qiming Yuan, Rachel Dias, Rachel Lim, Rahul Arora, Rajan Troll, Randall Lin, Rapha~Gontijo Lopes, Raul Puri, Reah Miyara, Reimar Leike, Renaud Gaubert, Reza Zamani, Ricky Wang, Rob Donnelly, Rob Honsby, Rocky Smith, Rohan Sahai, Rohit Ramchandani, Romain Huet, Rory Carmichael, Rowan Zellers, Roy Chen, Ruby Chen, Ruslan Nigmatullin, Ryan Cheu, Saachi Jain, Sam Altman, Sam Schoenholz, Sam
  Toizer, Samuel Miserendino, Sandhini Agarwal, Sara Culver, Scott Ethersmith, Scott Gray, Sean Grove, Sean Metzger, Shamez Hermani, Shantanu Jain, Shengjia Zhao, Sherwin Wu, Shino Jomoto, Shirong Wu, Shuaiqi, Xia, Sonia Phene, Spencer Papay, Srinivas Narayanan, Steve Coffey, Steve Lee, Stewart Hall, Suchir Balaji, Tal Broda, Tal Stramer, Tao Xu, Tarun Gogineni, Taya Christianson, Ted Sanders, Tejal Patwardhan, Thomas Cunninghman, Thomas Degry, Thomas Dimson, Thomas Raoux, Thomas Shadwell, Tianhao Zheng, Todd Underwood, Todor Markov, Toki Sherbakov, Tom Rubin, Tom Stasi, Tomer Kaftan, Tristan Heywood, Troy Peterson, Tyce Walters, Tyna Eloundou, Valerie Qi, Veit Moeller, Vinnie Monaco, Vishal Kuo, Vlad Fomenko, Wayne Chang, Weiyi Zheng, Wenda Zhou, Wesam Manassra, Will Sheu, Wojciech Zaremba, Yash Patil, Yilei Qian, Yongjik Kim, Youlong Cheng, Yu~Zhang, Yuchen He, Yuchen Zhang, Yujia Jin, Yunxing Dai, and Yury Malkov. 2024.
\newblock \href {https://arxiv.org/abs/2410.21276} {Gpt-4o system card}.
\newblock \emph{Preprint}, arXiv:2410.21276.

\bibitem[{Qwen et~al.(2025)Qwen, :, Yang, Yang, Zhang, Hui, Zheng, Yu, Li, Liu, Huang, Wei, Lin, Yang, Tu, Zhang, Yang, Yang, Zhou, Lin, Dang, Lu, Bao, Yang, Yu, Li, Xue, Zhang, Zhu, Men, Lin, Li, Tang, Xia, Ren, Ren, Fan, Su, Zhang, Wan, Liu, Cui, Zhang, and Qiu}]{qwen2025qwen25technicalreport}
Qwen, :, An~Yang, Baosong Yang, Beichen Zhang, Binyuan Hui, Bo~Zheng, Bowen Yu, Chengyuan Li, Dayiheng Liu, Fei Huang, Haoran Wei, Huan Lin, Jian Yang, Jianhong Tu, Jianwei Zhang, Jianxin Yang, Jiaxi Yang, Jingren Zhou, Junyang Lin, Kai Dang, Keming Lu, Keqin Bao, Kexin Yang, Le~Yu, Mei Li, Mingfeng Xue, Pei Zhang, Qin Zhu, Rui Men, Runji Lin, Tianhao Li, Tianyi Tang, Tingyu Xia, Xingzhang Ren, Xuancheng Ren, Yang Fan, Yang Su, Yichang Zhang, Yu~Wan, Yuqiong Liu, Zeyu Cui, Zhenru Zhang, and Zihan Qiu. 2025.
\newblock \href {https://arxiv.org/abs/2412.15115} {Qwen2.5 technical report}.
\newblock \emph{Preprint}, arXiv:2412.15115.

\bibitem[{Reimers and Gurevych(2019)}]{reimers-gurevych-2019-sentence}
Nils Reimers and Iryna Gurevych. 2019.
\newblock \href {https://doi.org/10.18653/v1/D19-1410} {Sentence-{BERT}: Sentence embeddings using {S}iamese {BERT}-networks}.
\newblock In \emph{Proceedings of the 2019 Conference on Empirical Methods in Natural Language Processing and the 9th International Joint Conference on Natural Language Processing (EMNLP-IJCNLP)}, pages 3982--3992, Hong Kong, China. Association for Computational Linguistics.

\bibitem[{Shaier et~al.(2024)Shaier, Kobren, and Ogren}]{shaier-etal-2024-adaptive}
Sagi Shaier, Ari Kobren, and Philip~V. Ogren. 2024.
\newblock \href {https://doi.org/10.18653/v1/2024.emnlp-main.956} {Adaptive question answering: Enhancing language model proficiency for addressing knowledge conflicts with source citations}.
\newblock In \emph{Proceedings of the 2024 Conference on Empirical Methods in Natural Language Processing}, pages 17226--17239, Miami, Florida, USA. Association for Computational Linguistics.

\bibitem[{Shi et~al.(2024)Shi, Han, Lewis, Tsvetkov, Zettlemoyer, and Yih}]{shi-etal-2024-trusting}
Weijia Shi, Xiaochuang Han, Mike Lewis, Yulia Tsvetkov, Luke Zettlemoyer, and Wen-tau Yih. 2024.
\newblock \href {https://doi.org/10.18653/v1/2024.naacl-short.69} {Trusting your evidence: Hallucinate less with context-aware decoding}.
\newblock In \emph{Proceedings of the 2024 Conference of the North American Chapter of the Association for Computational Linguistics: Human Language Technologies (Volume 2: Short Papers)}, pages 783--791, Mexico City, Mexico. Association for Computational Linguistics.

\bibitem[{Wang et~al.(2024{\natexlab{a}})Wang, Wan, Sun, Chen, and Arık}]{wang2024astuteragovercomingimperfect}
Fei Wang, Xingchen Wan, Ruoxi Sun, Jiefeng Chen, and Sercan~Ö. Arık. 2024{\natexlab{a}}.
\newblock \href {https://arxiv.org/abs/2410.07176} {Astute rag: Overcoming imperfect retrieval augmentation and knowledge conflicts for large language models}.
\newblock \emph{Preprint}, arXiv:2410.07176.

\bibitem[{Wang et~al.(2024{\natexlab{b}})Wang, Ren, Li, Zhao, Liu, and Wen}]{wang-etal-2024-rear}
Yuhao Wang, Ruiyang Ren, Junyi Li, Xin Zhao, Jing Liu, and Ji-Rong Wen. 2024{\natexlab{b}}.
\newblock \href {https://doi.org/10.18653/v1/2024.emnlp-main.321} {{REAR}: A relevance-aware retrieval-augmented framework for open-domain question answering}.
\newblock In \emph{Proceedings of the 2024 Conference on Empirical Methods in Natural Language Processing}, pages 5613--5626, Miami, Florida, USA. Association for Computational Linguistics.

\bibitem[{Wang et~al.(2024{\natexlab{c}})Wang, Li, Jiang, Tu, and Shi}]{wang-etal-2024-crafting}
Zheng Wang, Zhongyang Li, Zeren Jiang, Dandan Tu, and Wei Shi. 2024{\natexlab{c}}.
\newblock \href {https://doi.org/10.18653/v1/2024.emnlp-main.281} {Crafting personalized agents through retrieval-augmented generation on editable memory graphs}.
\newblock In \emph{Proceedings of the 2024 Conference on Empirical Methods in Natural Language Processing}, pages 4891--4906, Miami, Florida, USA. Association for Computational Linguistics.

\bibitem[{Wei et~al.(2024)Wei, Chen, and Meng}]{wei2024instructraginstructingretrievalaugmentedgeneration}
Zhepei Wei, Wei-Lin Chen, and Yu~Meng. 2024.
\newblock \href {https://arxiv.org/abs/2406.13629} {Instructrag: Instructing retrieval-augmented generation via self-synthesized rationales}.
\newblock \emph{Preprint}, arXiv:2406.13629.

\bibitem[{Xu et~al.(2024{\natexlab{a}})Xu, Lin, Yang, Zhang, Shi, Zhang, Fang, Xu, and Qiu}]{xu-etal-2024-earth}
Rongwu Xu, Brian Lin, Shujian Yang, Tianqi Zhang, Weiyan Shi, Tianwei Zhang, Zhixuan Fang, Wei Xu, and Han Qiu. 2024{\natexlab{a}}.
\newblock \href {https://doi.org/10.18653/v1/2024.acl-long.858} {The earth is flat because...: Investigating {LLM}s' belief towards misinformation via persuasive conversation}.
\newblock In \emph{Proceedings of the 62nd Annual Meeting of the Association for Computational Linguistics (Volume 1: Long Papers)}, pages 16259--16303, Bangkok, Thailand. Association for Computational Linguistics.

\bibitem[{Xu et~al.(2024{\natexlab{b}})Xu, Qi, Guo, Wang, Wang, Zhang, and Xu}]{xu-etal-2024-knowledge-conflicts}
Rongwu Xu, Zehan Qi, Zhijiang Guo, Cunxiang Wang, Hongru Wang, Yue Zhang, and Wei Xu. 2024{\natexlab{b}}.
\newblock \href {https://doi.org/10.18653/v1/2024.emnlp-main.486} {Knowledge conflicts for {LLM}s: A survey}.
\newblock In \emph{Proceedings of the 2024 Conference on Empirical Methods in Natural Language Processing}, pages 8541--8565, Miami, Florida, USA. Association for Computational Linguistics.

\bibitem[{Yang et~al.(2018)Yang, Qi, Zhang, Bengio, Cohen, Salakhutdinov, and Manning}]{yang-etal-2018-hotpotqa}
Zhilin Yang, Peng Qi, Saizheng Zhang, Yoshua Bengio, William Cohen, Ruslan Salakhutdinov, and Christopher~D. Manning. 2018.
\newblock \href {https://doi.org/10.18653/v1/D18-1259} {{H}otpot{QA}: A dataset for diverse, explainable multi-hop question answering}.
\newblock In \emph{Proceedings of the 2018 Conference on Empirical Methods in Natural Language Processing}, pages 2369--2380, Brussels, Belgium. Association for Computational Linguistics.

\bibitem[{Yu et~al.(2024)Yu, Gan, Zhang, Tong, Liu, and Liu}]{yu2024evaluationretrievalaugmentedgenerationsurvey}
Hao Yu, Aoran Gan, Kai Zhang, Shiwei Tong, Qi~Liu, and Zhaofeng Liu. 2024.
\newblock \href {https://arxiv.org/abs/2405.07437} {Evaluation of retrieval-augmented generation: A survey}.
\newblock \emph{Preprint}, arXiv:2405.07437.

\bibitem[{Zhou et~al.(2023)Zhou, Zhang, Poon, and Chen}]{zhou-etal-2023-context}
Wenxuan Zhou, Sheng Zhang, Hoifung Poon, and Muhao Chen. 2023.
\newblock \href {https://doi.org/10.18653/v1/2023.findings-emnlp.968} {Context-faithful prompting for large language models}.
\newblock In \emph{Findings of the Association for Computational Linguistics: EMNLP 2023}, pages 14544--14556, Singapore. Association for Computational Linguistics.

\bibitem[{Zhu et~al.(2024)Zhu, Luo, Xu, Wang, Yu, Wang, Yan, Liu, Han, Liu, and Sun}]{zhu2024ragevalscenariospecificrag}
Kunlun Zhu, Yifan Luo, Dingling Xu, Ruobing Wang, Shi Yu, Shuo Wang, Yukun Yan, Zhenghao Liu, Xu~Han, Zhiyuan Liu, and Maosong Sun. 2024.
\newblock \href {https://arxiv.org/abs/2408.01262} {Rageval: Scenario specific rag evaluation dataset generation framework}.
\newblock \emph{Preprint}, arXiv:2408.01262.

\end{thebibliography}

\appendix

\section{Detailed Dataset Curation Procedure}
\label{sec:dataset_construction}

Below, we provide a step-by-step description of how we constructed the URAQ dataset:

\subsection{Knowledge Generation}
We used \emph{gpt-4o-mini}~\cite{openai2024gpt4ocard} to produce an initial list of short, simple knowledge statements. These statements are general facts (e.g., ``A hummingbird can hover in mid-air'' or ``Blue whales are the largest animals on Earth'') rather than domain-specific or specialized knowledge. The generated statements were deliberately kept concise and straightforward to facilitate subsequent manipulation and question generation.

\subsection{Redundancy Filtering}
Since GPT-based generators can produce highly similar or paraphrased statements, we employed \emph{SentenceBERT}~\cite{reimers-gurevych-2019-sentence} to measure the semantic similarity between all knowledge statements. Any pair of statements with a cosine similarity above 0.5 was considered near-duplicate and therefore removed to ensure diversity in the final knowledge set.

\subsection{Manipulated Knowledge Creation}
For every remaining ``original'' knowledge statement, we prompted \emph{gpt-4o-mini} to generate a \emph{manipulated} variant. The manipulation involved either substituting key elements (e.g., entities, numerical values, or critical details) or adding a negation that changes the statement's truth value (e.g., ``A hummingbird cannot hover in mid-air''). Each pair of statements (original vs.\ manipulated) thus serves as a pairwise contrast for subsequent question-answer (QA) creation.

\subsection{Question-Answer (QA) Generation}
From each pair of original and manipulated knowledge statements, we prompted \emph{gpt-4o-mini} to generate a question that requires between 1 to 5 \emph{reasoning steps} to arrive at an answer. The reasoning steps typically involve either numerical computation, logical inference, or entity comparison. Each question was tied to both the original and the manipulated knowledge. The resulting QA format consists of one question and two different answers: one correct answer derived from the original statement, and a second answer derived from the manipulated statement.

\subsection{Answer Format and Difficulty Selection}
We constrained valid answers to be either (i) a numeric value, (ii) a boolean (``yes'' or ``no''), or (iii) a single entity. Among the generated questions, those requiring 4-hop reasoning were chosen for the final dataset, as manual inspection suggested these exhibited higher quality and clearer multi-step logic compared to simpler or more complex variants.

\subsection{Final Ground Truth Assignment}
For each question, we designated the correct ground truth answer to be the one aligned with the \emph{original} knowledge statement. An example illustrating how this ground truth is integrated into the evaluation framework is provided in Figure~\ref{fig:framework} of the main paper.

By following these steps, we ensure that the URAQ dataset offers well-defined pairs of knowledge (original vs.\ manipulated) and corresponding multi-step questions designed to differentiate between factual and altered information. This framework supports a diverse range of potential use cases, from fact-checking systems to more elaborate multi-step reasoning models.

\section{Example User Need Instructions}
\label{sec:example_user_need}
\subsection{Context-Exclusive}
\begin{lstlisting}[language=Python, style=jsonstyle]
You are a helpful AI assistant tasked with answering the given question ONLY based on the provided information. Here are the requirements to answer the question:

1. The answer should be a numeric value, a boolean ("yes" or "no"), or an entity.

2. You MUST directly provide the final answer within an <output> XML tag, without including any units if the answer is numeric.

3. You MUST utilize the RELEVANT knowledge contained in the provided information to answer the question, even if the knowledge is INCORRECT. If NONE of the provided information is RELEVANT to the question, you MUST output "I don't know".
\end{lstlisting}

\subsection{Context-First}
\begin{lstlisting}[language=Python, style=jsonstyle]
You are a helpful AI assistant tasked with answering the given question by referring to the provided information. Here are the requirements to answer the question:

1. The answer should be a numeric value, a boolean ("yes" or "no"), or an entity.

2. You MUST directly provide the final answer within an <output> XML tag, without including any units if the answer is numeric.

3. If the provided information contains RELEVANT knowledge that can be used to answer the question, you MUST utilize the provided information, even if the knowledge is INCORRECT.

4. If NONE of the provided information is RELEVANT to the question, you MUST utilize your own knowledge to answer the question.
\end{lstlisting}

\subsection{Memory-First}
\begin{lstlisting}[language=Python, style=jsonstyle]
You are a helpful AI assistant tasked with answering the given question by referring to the provided information. Here are the requirements to answer the question:

1. The answer should be a numeric value, a boolean ("yes" or "no"), or an entity.

2. You MUST directly provide the final answer within an <output> XML tag, without including any units if the answer is numeric.

3. You MUST utilize your own knowledge to answer the question if you are certain of the accuracy (e.g., factual information you are sure about). If you are UNSURE about your knowledge, you MUST use the relevant knowledge from the given information instead.
\end{lstlisting}

\section{Input Prompt Formatting}
\label{sec:example_input_prompt}
The input prompt is organized as $(I, C, Q)$ or $(I_f, I_u, C, Q)$, where $I$ is the instruction and can be separated into formatting instruction $I_f$ and user needs instructions $I_u$, $C=\{c_1, c_2, ..., c_n\}$ is a series of retrieved context with retrieval number of $n$, and $Q$ is the question. Given an input $(I_f, I_u, C, Q)$, we have the following prompting template: 
\begin{equation}
    \text{<sys>}I_f\oplus I_u\text{</sys>}\text{<user>}C\oplus Q\text{</user>}
\end{equation}
where <sys></sys> and <user></user> denote the system prompt and the user prompt. Among all data samples, the $I_u$ and $C$ may change according to the \textbf{user case} and \textbf{context setting}, while the $I_f$ remaining the same by instructing models to directly output a simple answer that is either a numeric value, a boolean ("yes" or "no"), or an entity, as described in Section \ref{subsec:datasets}.

We introduce an example input prompt that is designed for \textbf{Context-Exclusive and Context Matching} with 2 total retrieved context following the abstract input $(I_f, I_u, C, Q)$. The prompt is formatted with XML for both input and output. Specifically, the formatting instructions $I_f$ are separated into two parts: 1) The first and second instructions in the system prompt describing that the answer should be as simple as possible with XML format. 2) The instruction in the user prompt about format of context with an reinforcement of output format. The \emph{user need} instruction $I_u$ is at the third instruction in the system prompt. The retrieved context $C$ is all the sentences in the user prompt within <info> xml tag. Finally, the question $Q$ is at the last of user prompt within <question> xml tag. 
\begin{lstlisting}[language=Python, style=jsonstyle]
System Prompt:
You are a helpful AI assistant tasked with answering the given question ONLY based on the provided information. Here are the requirements to answer the question:
1. The answer should be a numeric value, a boolean ("yes" or "no"), or an entity.
2. You MUST directly provide the final answer within an <output> XML tag, without including any units if the answer is numeric.
3. You MUST utilize the RELEVANT knowledge contained in the provided information to answer the question, even if the knowledge is INCORRECT. If NONE of the provided information is RELEVANT to the question, you MUST output 'I don't know'.

User Prompt:
Given the following information enclosed in xml tag <info></info>, answer the question enclosed in xml tag <question></question>.
**
<info>2 + 2 equals 4.</info>
<info>The sun is the only star in our solar system.</info>
**
<question>If you have two apples and you get two more, how many apples do you have?</question>
\end{lstlisting}

\section{Case Study of Model Laziness}
\label{sec:case_study_of_model_lazy}
\begin{lstlisting}[language=Python, style=jsonstyle]
Question: How do plants create their food, photosynthesis or moonthesis?

Answer with 1 context: <output>No</output>

Answer with 1000 context: I think there might be a small mistake in your question. The correct term is "photosynthesis" not "moonthesis". Photosynthesis is the process by which plants, algae, and some bacteria make their own food using sunlight, water, and carbon dioxide. So, to answer your question: Plants create their food through photosynthesis.
\end{lstlisting}
Above question-answer pair from our synthetic URAQ dataset is an example of model relying on their own memory on long context and acquire better result than the result from shorter input. For answer with only 1 context, the model outputs a single word "No" with correct output format. For answer with 1000 context, the model provides clear thinking path with correct, affirmative answer without the desinated XML format for output. We also calculate the percentage of 100 randomly selected testing samples that has similar behaviors using Qwen2.5-72B-Instruct and Llama-3.1-70B-Instruct as shown in Table \ref{tab:model_lazy}. 

\begin{table}[htbp!]
\centering\resizebox{\columnwidth}{!}{
\begin{tabular}{c|cc} 
\toprule
 & \bf Context-First (\%) & \bf Memory-First (\%) \\
\midrule
Qwen2.5-72B-Instruct & 84 & 77 \\
Llama-3.1-70B-Instruct & 56 & 65 \\
\bottomrule
\end{tabular}}
\caption{Percentage of testing samples that answered with single negative output for short input but correct output with explicit reasoning, among 100 randomly selected samples that the question answered incorrectly with 1 retrieved context and correctly with 1000 retrieved context.}
\label{tab:model_lazy}
\end{table}

\section{Accuracy Curves of URAQ and DisentQA}
\label{sec:accuracy_curve_of_all_datasets}

\begin{figure}[htbp!]
\centering

\begin{subfigure}{\columnwidth}
\centering
\includegraphics[width=\columnwidth]{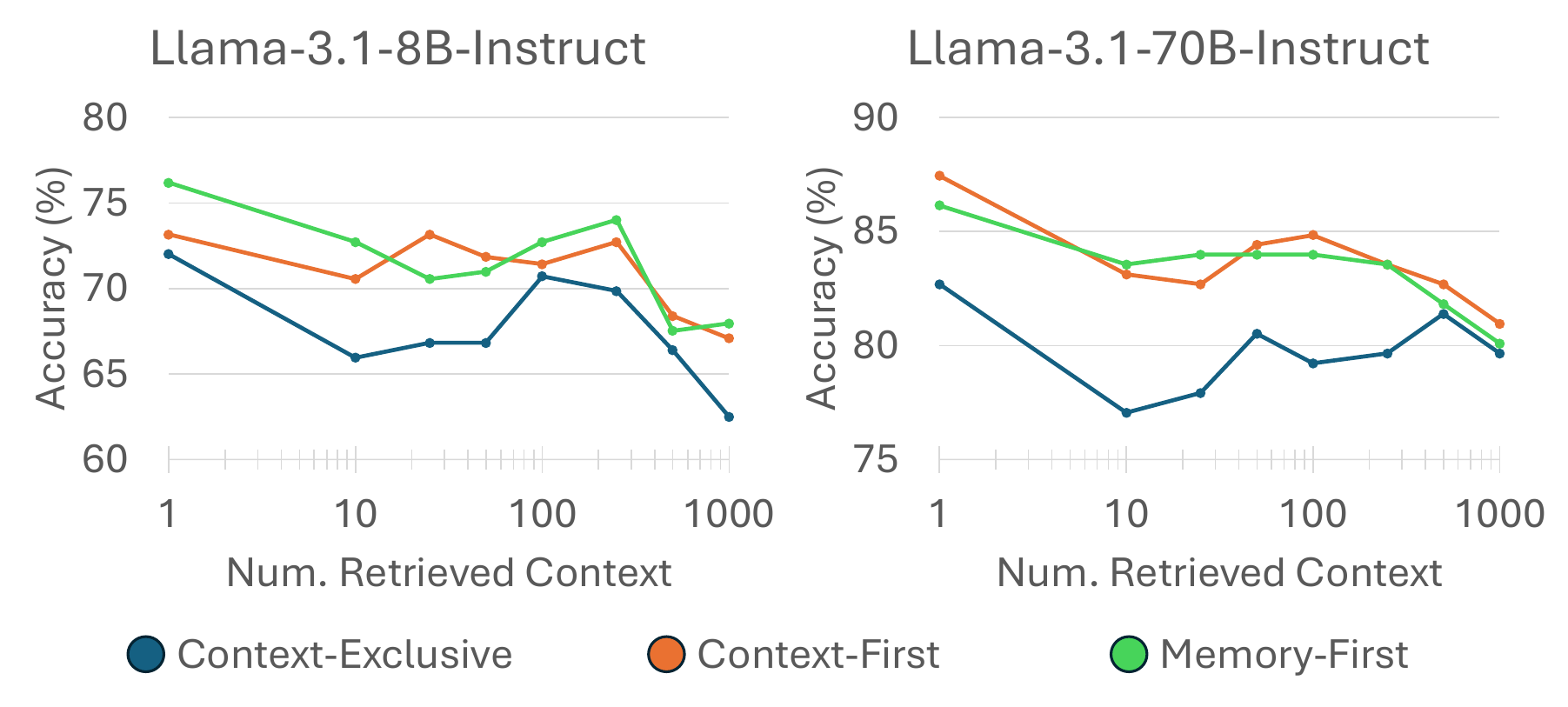}
\caption{Accuracy curve of Llama-3.1 on URAQ dataset under \emph{Context Matching} setting.}
\end{subfigure}

\begin{subfigure}{\columnwidth}
\centering
\includegraphics[width=\columnwidth]{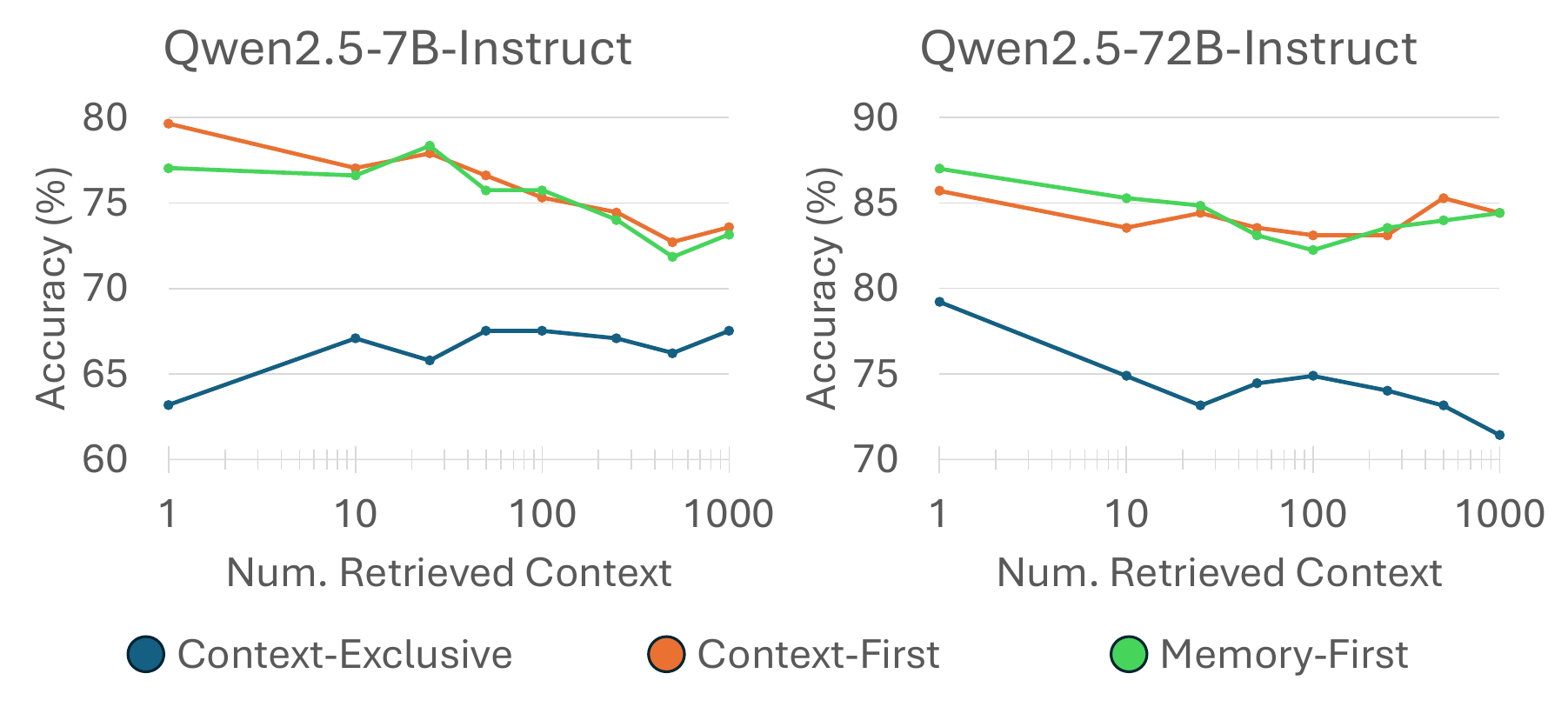}
\caption{Accuracy curve of Qwen2.5 on URAQ dataset under \emph{Context Matching} setting.}
\end{subfigure}

\begin{subfigure}{\columnwidth}
\centering
\includegraphics[width=\columnwidth]{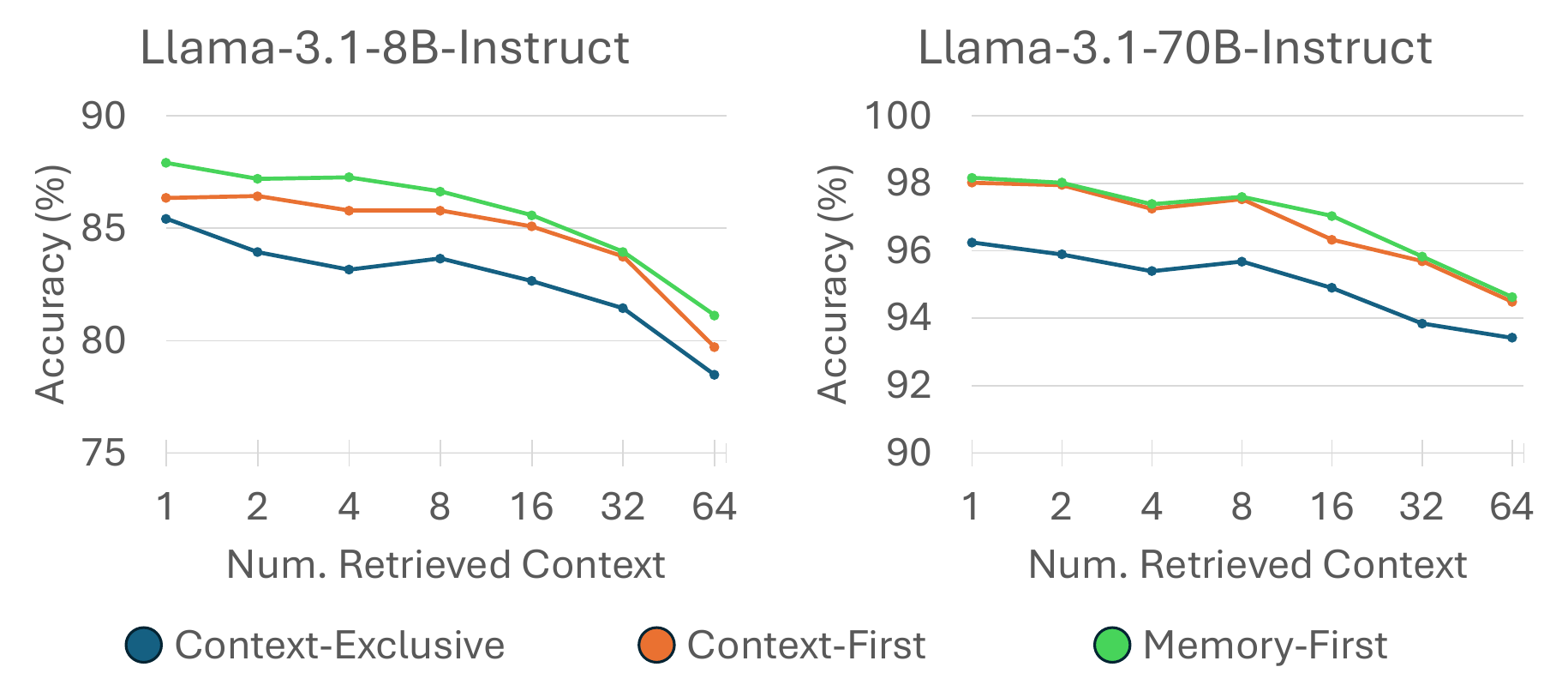}
\caption{Accuracy curve of Llama-3.1 on DisentQA dataset under \emph{Context Matching} setting.}
\end{subfigure}

\begin{subfigure}{\columnwidth}
\centering
\includegraphics[width=\columnwidth]{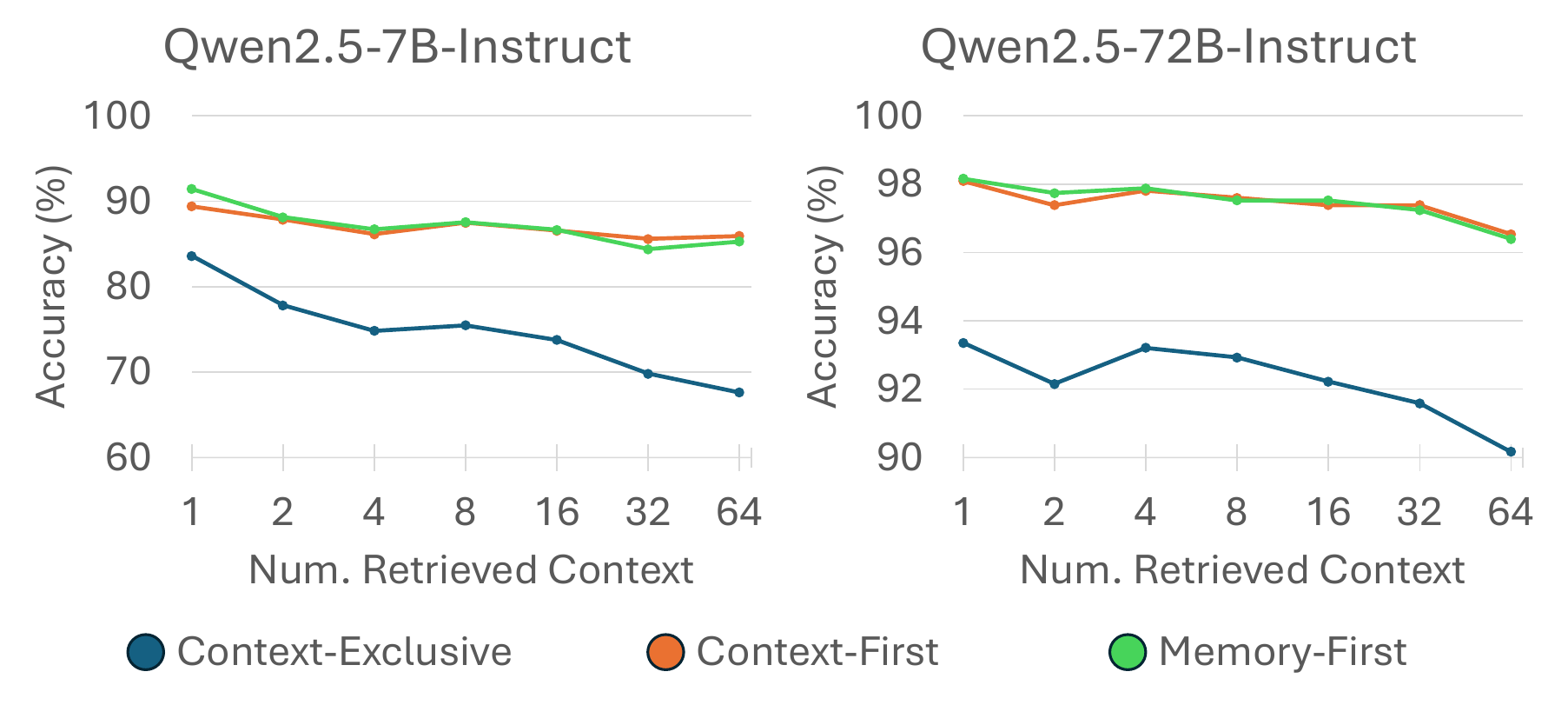}
\caption{Accuracy curve of Qwen2.5 on DisentQA dataset under \emph{Context Matching} setting.}
\end{subfigure}

\caption{Accuracy curve of all models under \emph{Context Matching} setting.}
\label{fig:setting_a_total}
\end{figure}


\begin{figure}[htbp!]
\centering

\begin{subfigure}{\columnwidth}
\centering
\includegraphics[width=\columnwidth]{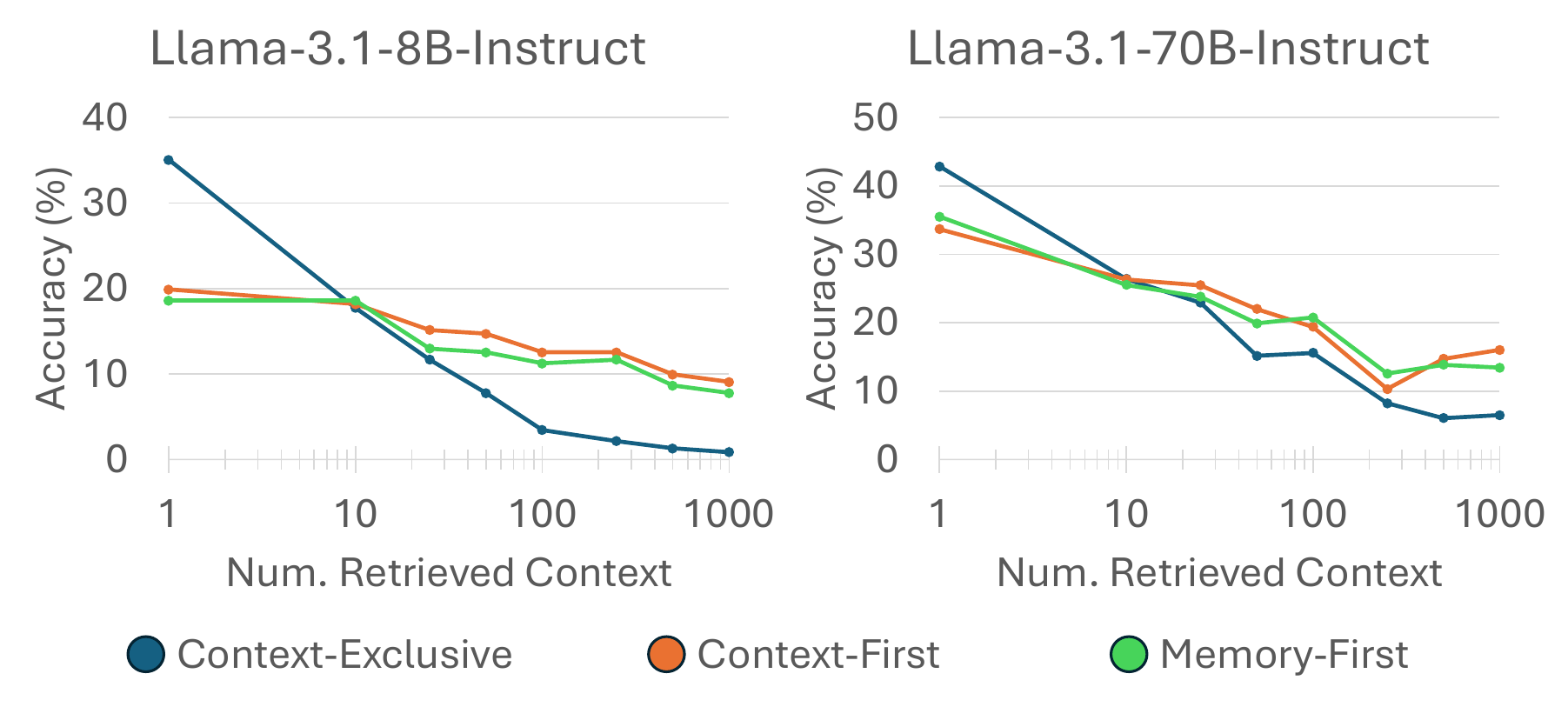}
\caption{Case-Level Accuracy curve of Llama-3.1 on URAQ dataset.}
\end{subfigure}

\begin{subfigure}{\columnwidth}
\centering
\includegraphics[width=\columnwidth]{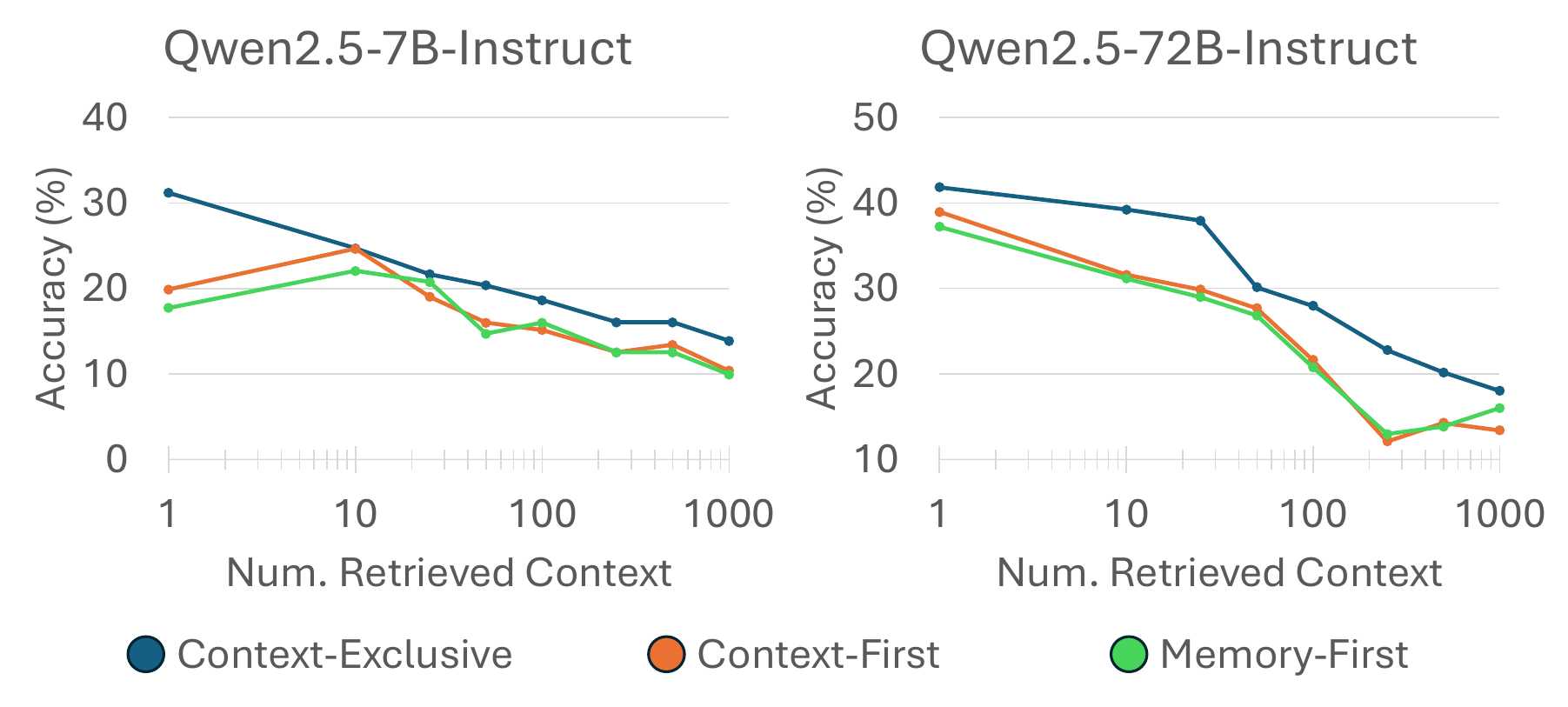}
\caption{Case-Level Accuracy curve of Qwen2.5 on URAQ dataset.}
\end{subfigure}

\begin{subfigure}{\columnwidth}
\centering
\includegraphics[width=\columnwidth]{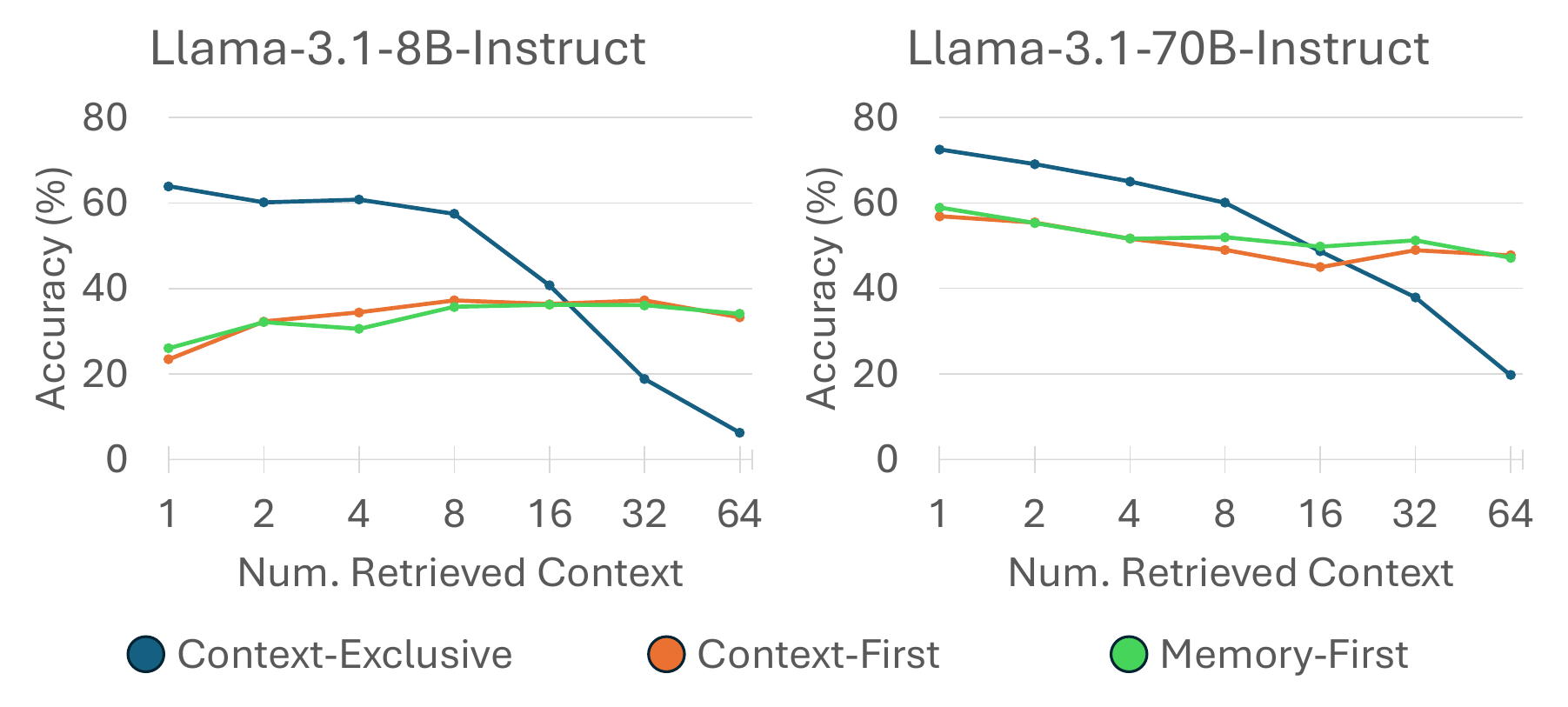}
\caption{Case-Level Accuracy curve of Llama-3.1 on DisentQA dataset.}
\end{subfigure}

\begin{subfigure}{\columnwidth}
\centering
\includegraphics[width=\columnwidth]{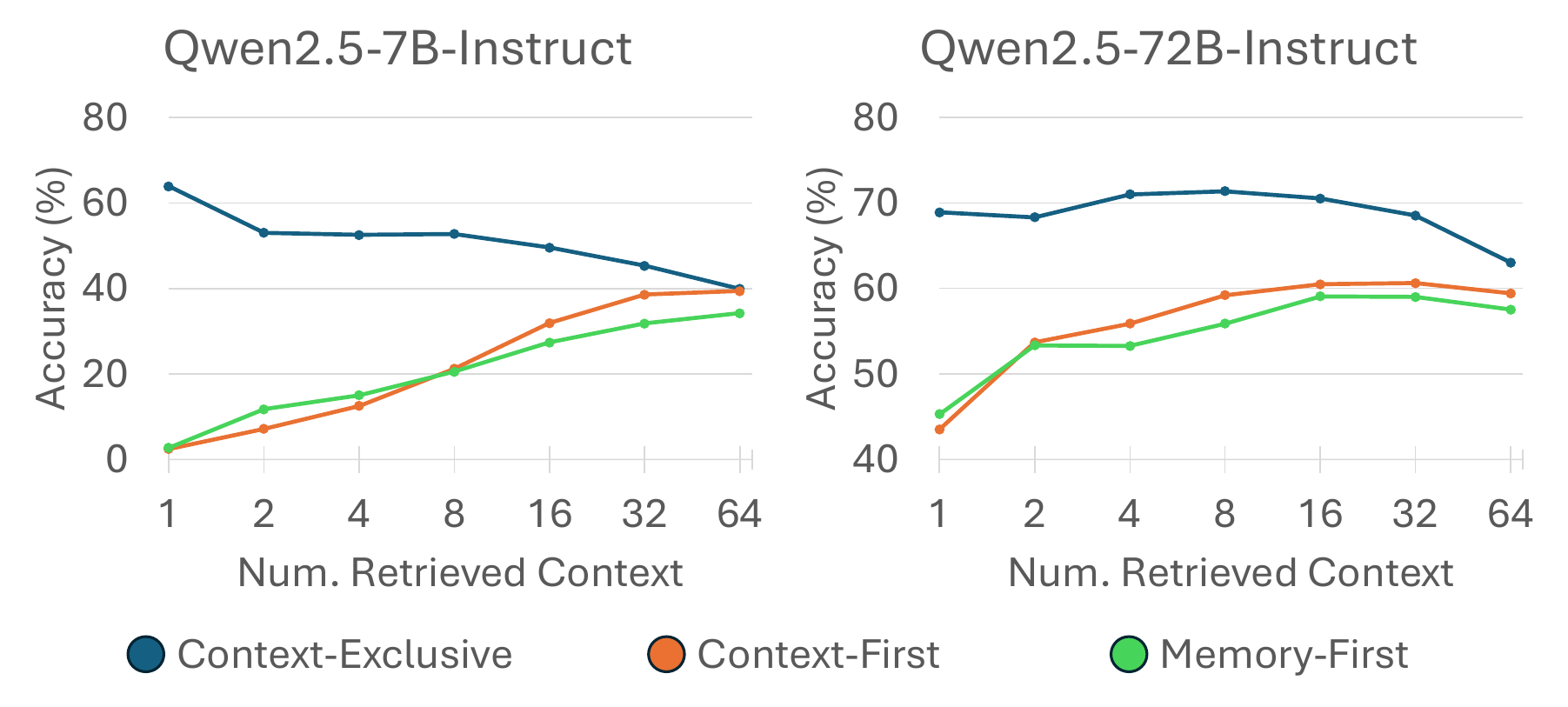}
\caption{Case-Level Accuracy curve of Qwen2.5 on DisentQA dataset.}
\end{subfigure}

\caption{Case-Level Accuracy of all models.}
\label{fig:setting_abc_total}
\end{figure}


\begin{figure}[htbp!]
\centering

\begin{subfigure}{\columnwidth}
\centering
\includegraphics[width=\columnwidth]{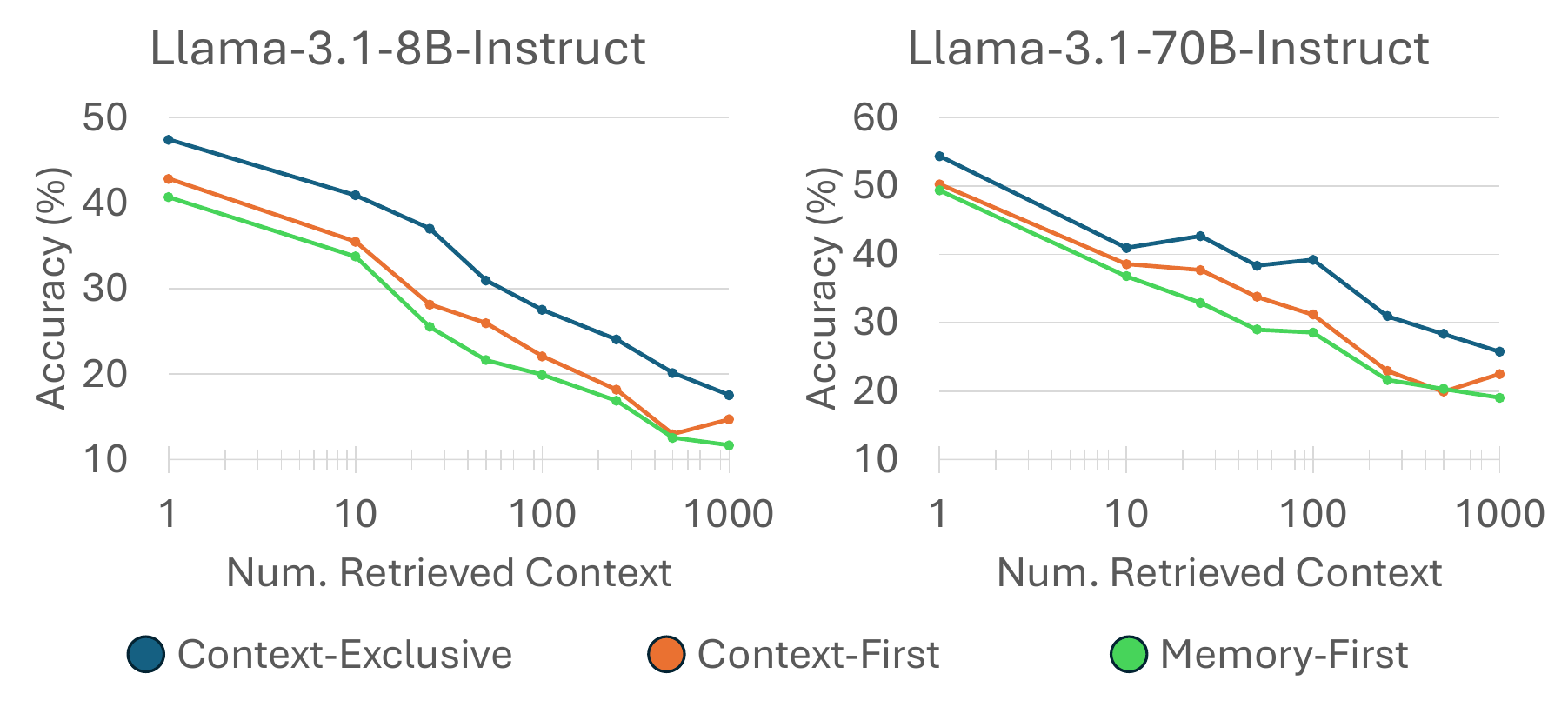}
\caption{Accuracy curve of Llama-3.1 on URAQ dataset under \emph{Context Matching \& Knowledge Conflict} setting.}
\end{subfigure}

\begin{subfigure}{\columnwidth}
\centering
\includegraphics[width=\columnwidth]{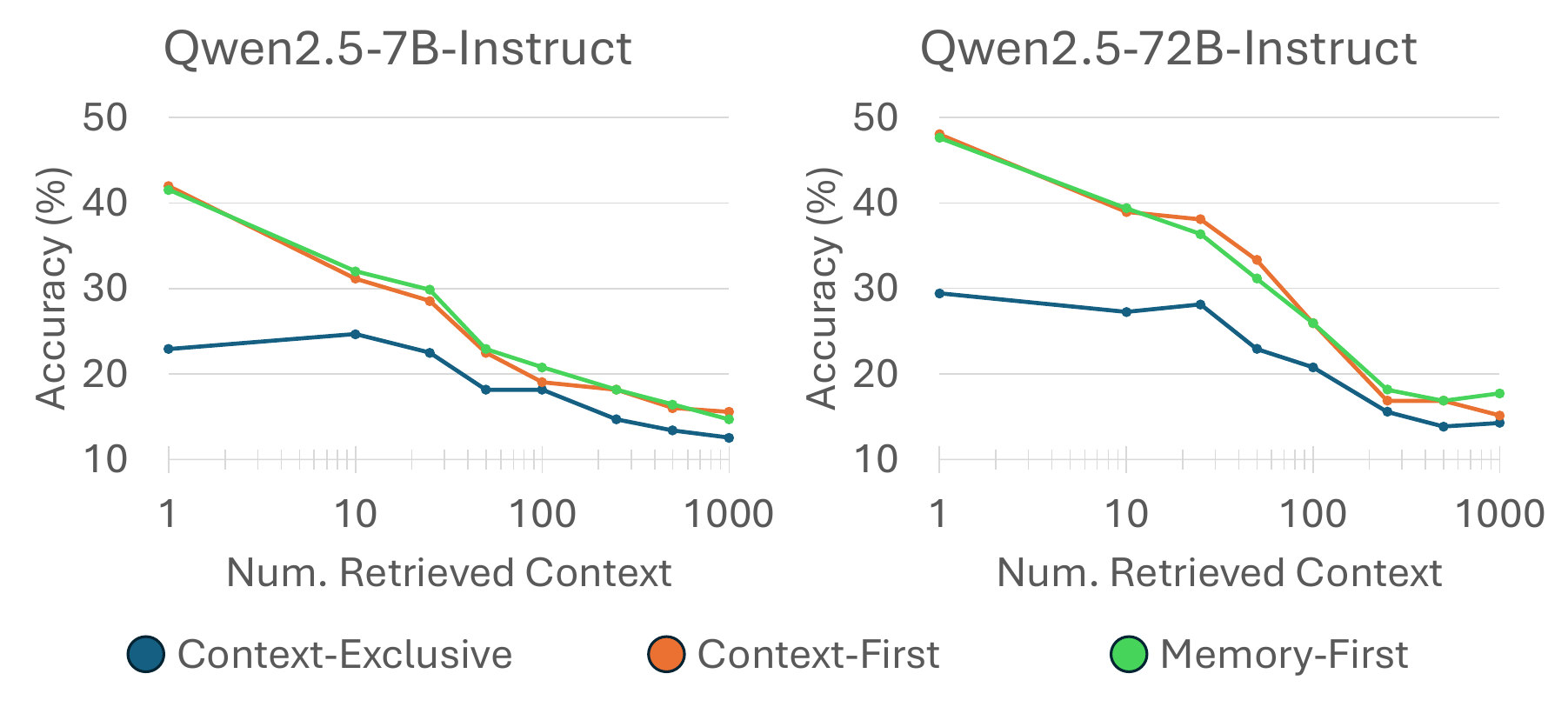}
\caption{Accuracy curve of Qwen2.5 on URAQ dataset under \emph{Context Matching \& Knowledge Conflict} setting.}
\end{subfigure}

\begin{subfigure}{\columnwidth}
\centering
\includegraphics[width=\columnwidth]{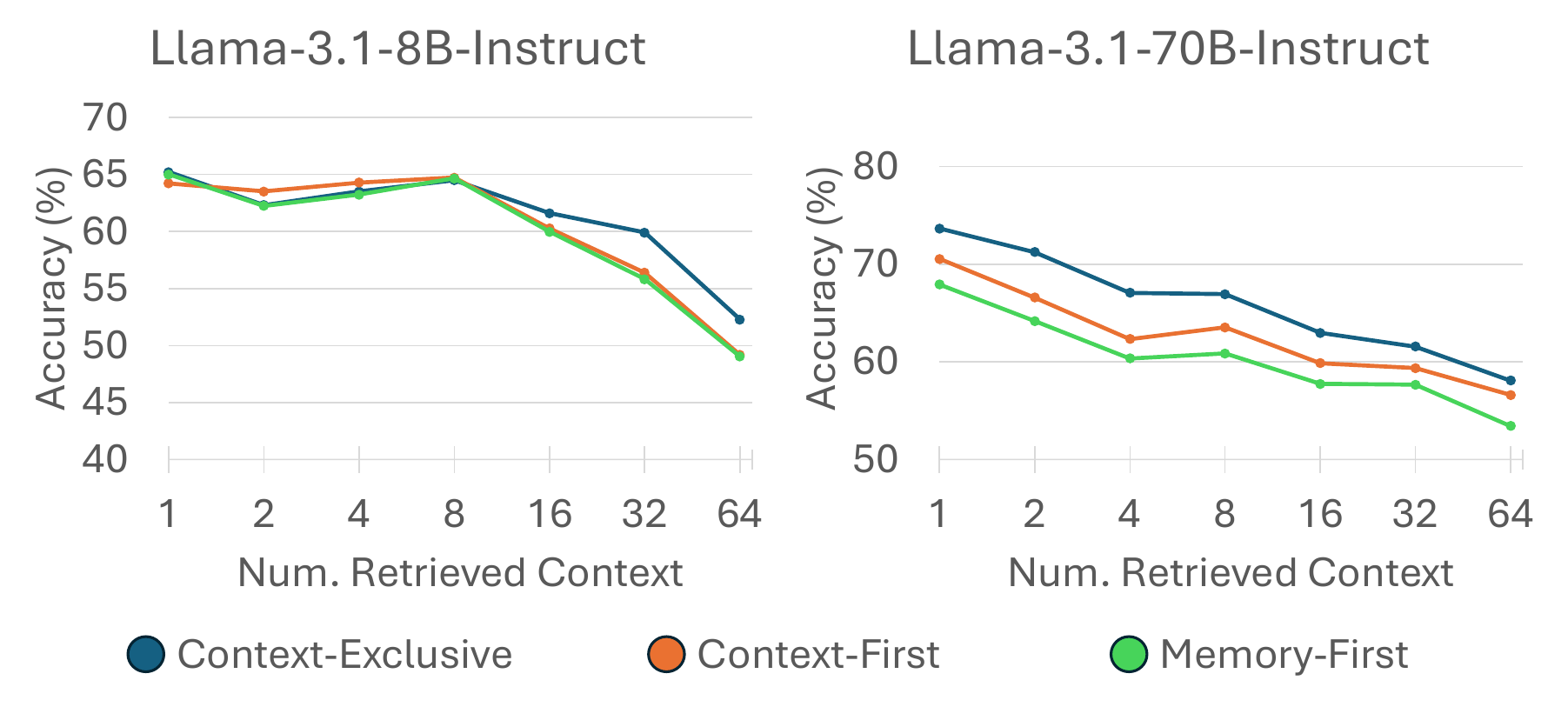}
\caption{Accuracy curve of Llama-3.1 on DisentQA dataset under \emph{Context Matching \& Knowledge Conflict} setting.}
\end{subfigure}

\begin{subfigure}{\columnwidth}
\centering
\includegraphics[width=\columnwidth]{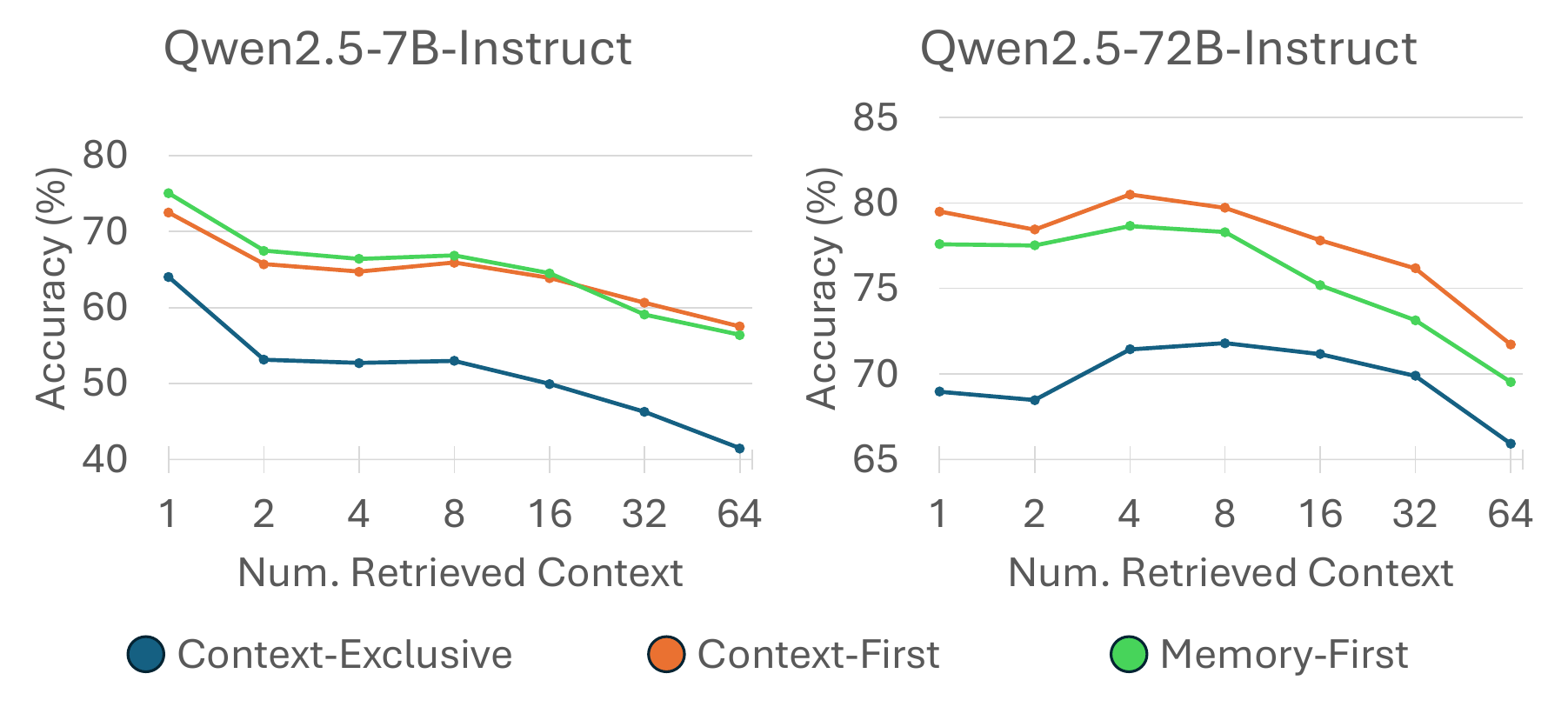}
\caption{Accuracy curve of Qwen2.5 on DisentQA dataset under \emph{Context Matching \& Knowledge Conflict} setting.}
\end{subfigure}

\caption{Accuracy curve of all models under \emph{Context Matching \& Knowledge Conflict} setting.}
\label{fig:setting_ab_total}
\end{figure}


\begin{figure}[htbp!]
\centering

\begin{subfigure}{\columnwidth}
\centering
\includegraphics[width=\columnwidth]{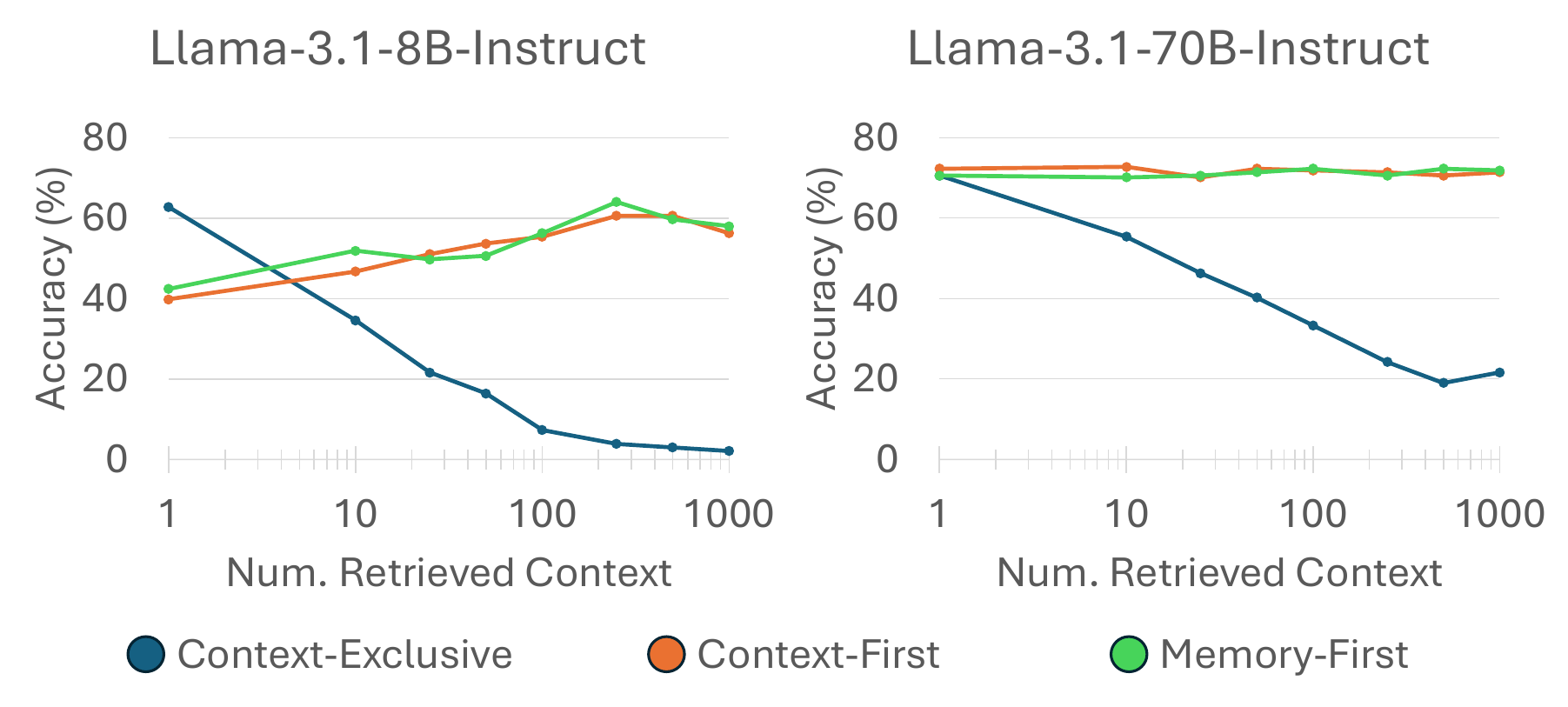}
\caption{Accuracy curve of Llama-3.1 on URAQ dataset under \emph{Context Matching \& Information Irrelevant} setting.}
\end{subfigure}

\begin{subfigure}{\columnwidth}
\centering
\includegraphics[width=\columnwidth]{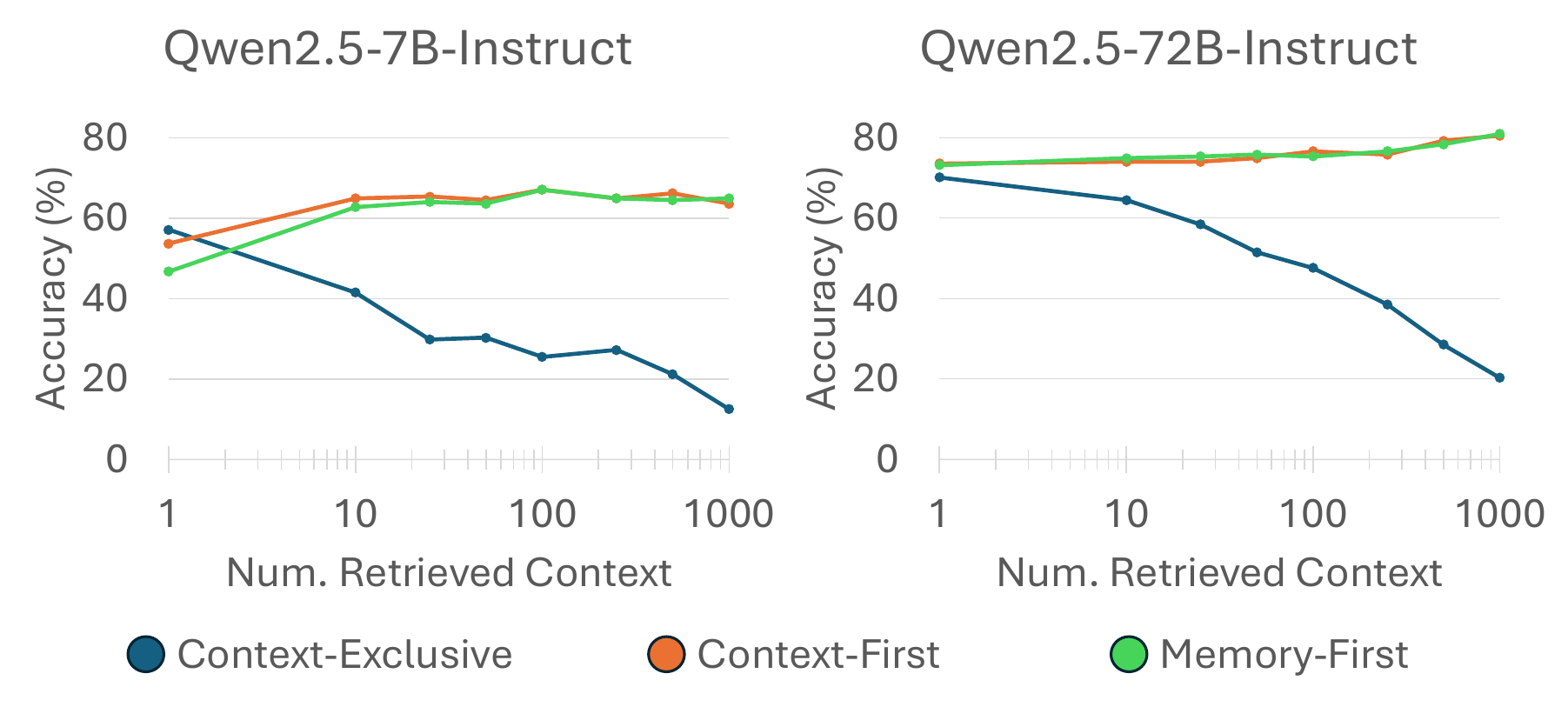}
\caption{Accuracy curve of Qwen2.5 on URAQ dataset under \emph{Context Matching \& Information Irrelevant} setting.}
\end{subfigure}

\begin{subfigure}{\columnwidth}
\centering
\includegraphics[width=\columnwidth]{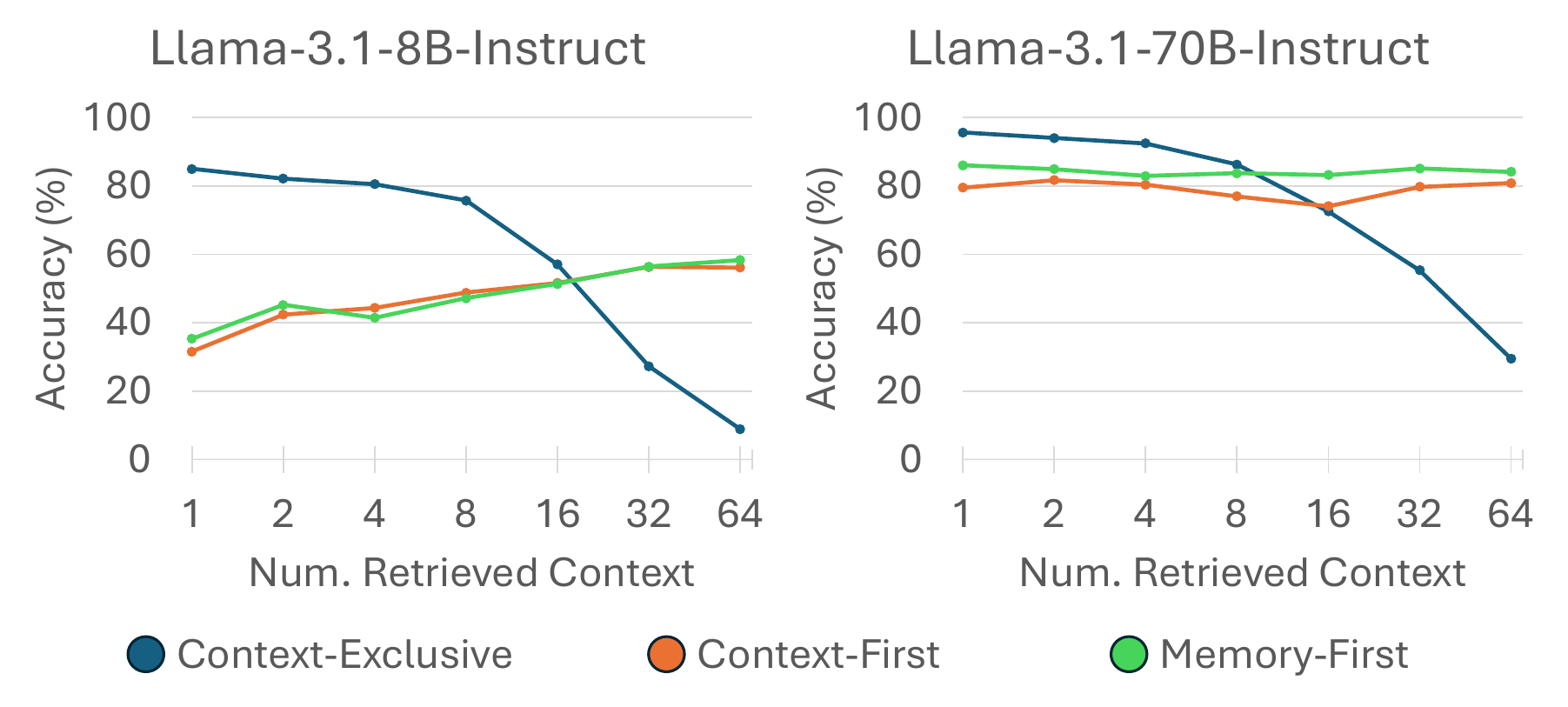}
\caption{Accuracy curve of Llama-3.1 on DisentQA dataset under \emph{Context Matching \& Information Irrelevant} setting.}
\end{subfigure}

\begin{subfigure}{\columnwidth}
\centering
\includegraphics[width=\columnwidth]{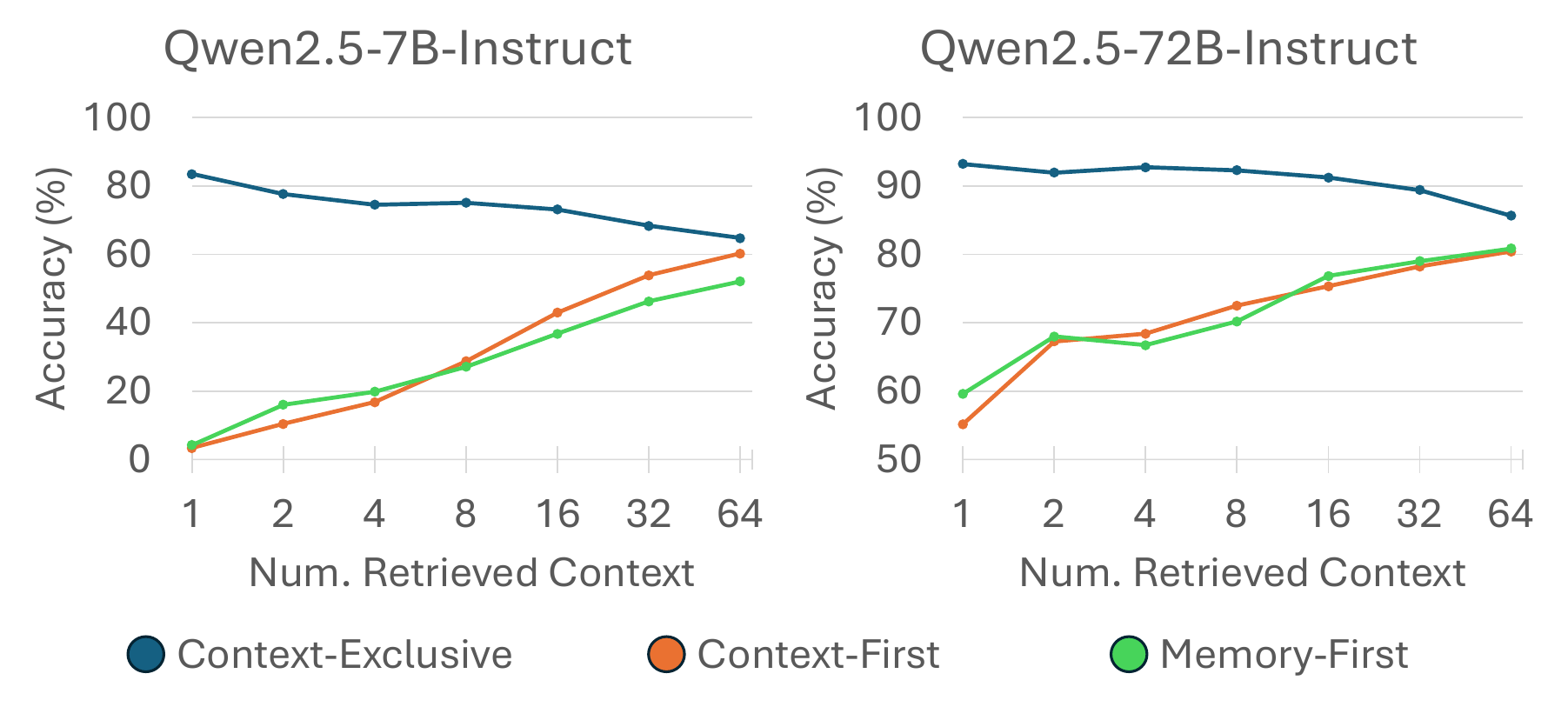}
\caption{Accuracy curve of Qwen2.5 on DisentQA dataset under \emph{Context Matching \& Information Irrelevant} setting.}
\end{subfigure}

\caption{Accuracy curve of all models under \emph{Context Matching \& Information Irrelevant} setting.}
\label{fig:setting_ac_total}
\end{figure}

\end{document}